\documentclass[12pt]{article} 

\pdfoutput=1

\usepackage{amsmath, amsfonts, amssymb}
\usepackage{titlesec}
\usepackage[margin=0.75in]{geometry}
\usepackage{graphicx}
\usepackage[hidelinks]{hyperref}
\usepackage{setspace}
\usepackage{algorithm}
\usepackage{algpseudocode}
\usepackage{tabularx}
\usepackage{lscape} 
\usepackage{rotating}
\usepackage{hyphenat} 
\usepackage{enumitem}
\usepackage{bm}
\usepackage{placeins} 
\usepackage{setspace}
\usepackage[capitalise]{cleveref}
\usepackage{fancyhdr}
\usepackage{xcolor}
\usepackage{adjustbox} 
\usepackage{tikz} 

\usepackage{longtable} 
\usepackage{caption}

\usepackage{multirow}

\usepackage[flushleft]{threeparttable}

\usepackage{booktabs}
\usepackage{multirow}

\usepackage{makecell} 

\usepackage{array, booktabs, makecell} 

\usepackage{float}

\def \bx{\mathbf{x}}

\usepackage{xcolor}

\definecolor{ForestGreen}{RGB}{34,139,34}

\newcommand{\edit}[1]{\color{black}#1 \color{black}}

\usepackage[backend=bibtex,style=numeric,giveninits=true,maxbibnames=99,maxcitenames=99,sorting=none]{biblatex}

\title{Self-explainable Operator Learning for Discovering Spatial Patterns in Functional Data}

\author{Mojgan Alishiri$^{1,2}$ \and Amirhossein Arzani$^{1,2}$ }

\date{}

\begin{document}

\maketitle

\begin{center}
$^1$Scientific Computing and Imaging Institute, University of Utah, Salt Lake City, UT, 84112, United States \\
$^2$Mechanical Engineering Department, University of Utah, Salt Lake City, UT, 84112, United States \\
\end{center}

\bigskip


\thispagestyle{empty}


\begin{abstract}
Operator learning has emerged as a powerful tool for modeling complex physical systems in functional spaces. However, their neural network–based architectures make them opaque models, obscuring the reasoning behind their predictions. In this work, we introduce a self-explainable operator learning framework that overcomes this challenge by reformulating operator learning as a linear combination of generalized functional linear models expressed through integral equations. Exploiting the additive decomposability of these integral equations, we divide the input domain into subdomains and compute localized integrals to evaluate the contribution of each region to the final prediction. This decomposition enables direct interpretability where the model explains both inputs and outputs by linking specific input regions to corresponding output patterns, thereby revealing which spatial features drive predictions. We demonstrate the framework on function-to-scalar and function-to-function mappings in fluid flow problems involving blood flow and unsteady aerodynamics. \edit{The results show that the operator most often prioritizes regions with strong feature gradients, providing physically meaningful insight into the model’s decision-making process. Comparisons with established post-hoc explainability methods demonstrate qualitative agreement while highlighting the key advantage of the proposed approach: explainability is embedded directly within the operator structure itself and does not require an external tool. Therefore, our framework provides a mathematically transparent and physically interpretable approach to uncover relationships within data, fostering trust in machine learning for scientific applications by enabling more informed data-driven analysis of physical systems.}
\end{abstract}
\noindent\textbf{Keywords:} Integral operators; Explainable Artificial Intelligence; Interpretability; Data-driven discovery; Unsteady fluid dynamics

\newpage

\pagenumbering{arabic}


\section{Introduction} \label{sec:intro}

Supervised machine learning (ML) models excel at predicting outcomes for test data within the distribution of the training data. For scientists to use and trust these predictions, it is essential not only to assess the accuracy of the model but also to ensure interpretability, allowing users to understand what the model has learned and verify its consistency with established scientific principles~\cite{rudin2019stop, ghiringhelli2021interpretability}. The added interpretability can position ML models as potential new scientific models rather than mere predictive surrogates~\cite{freiesleben2024scientific, muckley2023interpretable}. A primary challenge in discovering interpretable models for physical systems is inferring governing laws from data that represent discrete measurements of continuous, often high- or infinite-dimensional functions~\cite{kissas2022learning}. 

Operator learning in functional spaces offers a transformative approach to modeling complex physical systems, such as those governed by partial differential equations (PDEs), through architectures like neural operators, bridging ML techniques with functional analysis~\cite{kovachki2023neural}. This approach is particularly valuable for systems with nonlocal interactions, where a solution at any point depends on the input function across the entire domain, as exemplified by Green’s functions in PDEs~\cite{gu2025explainable}. Kernels naturally encode these interactions via integral operators. A notable challenge in functional analysis is accurately learning the kernel function that appears in integral operators~\cite{cao2022understand, kissas2022learning, li2023scalable,arzani2024interpreting}. This challenge has usually been addressed by expanding kernels in specific bases or utilizing pre-defined fixed kernels~\cite{horvath2012inference}, which may limit flexibility. Moreover, when the governing PDEs are unknown, operator learning serves as a surrogate to infer models from data or estimate unknown parameters. However, the trustworthiness of these predictions also depends on the interpretability of the operator itself, which poses a critical need for interpretability and explainability of this new deep learning paradigm.

Despite the efficiency of operator learning models like Fourier Neural Operators (FNO)~\cite{li2020fourier} and Deep Operator Network (DeepONet)~\cite{lu2019deeponet} as surrogates for physical systems, their complex architectures often render them opaque universal approximators, obscuring the reasoning behind their predictions. To ensure scientifically valid insights from these powerful yet opaque models, transparency is a desired feature~\cite{lipton2018mythos, samek2017explainable}. Transparency in such models can be assessed at three levels: whole-model transparency (simulatability), where a human can step through the input--output mapping to reproduce predictions; component transparency (decomposability), where structural choices such as kernel selection are explicitly justified through domain knowledge; and training-process transparency (algorithmic transparency), where the learning procedure is clear, reproducible, and free of hidden complexities~\cite{lipton2018mythos, hofmann2008kernel, roscher2020explainable}. Neural network--based operators, with their massive number of learnable parameters, fall short on all three counts, highlighting the need to address both interpretability and explainability. While the two terms are often used interchangeably under the umbrella of explainable AI (XAI)~\cite{murdoch2019interpretable, roscher2020explainable, samek2017explainable, marcinkevivcs2023interpretable}, they serve distinct roles: interpretability translates a model's internal reasoning into understandable terms --- for instance, elucidating how kernel functions influence outputs --- whereas explainability embeds those insights within domain-specific contexts to support model auditing, regulatory compliance, or new knowledge discovery~\cite{roscher2020explainable, von2021informed}.

Most opaque ML models, including neural operator architectures, rely on ``post-hoc'' methods to be explained. In contrast, interpretable ML prioritizes ``by-design'' models with inherently transparent structures and decision-making processes. By explaining the internal workings of their structure, interpretable approaches can achieve scientific explanations and potentially uncover novel insights~\cite{daniels2019automated, zappala2024learning, menier2025interpretable}. Self-interpretable models support simulatability where a human can follow the model's logic step-by-step, and their components are presented in a way that domain experts can interpret without additional processing. A range of interpretable model types have been developed to convey complex data relationships transparently~\cite{hanyu2022design, ustun2016supersparse, hastie1986generalized, ravikumar2009sparse, cranmer2023interpretable, marcinkevivcs2023interpretable}. However, in neural operator learning, the sequential addition of kernels across hidden layers compounds the opacity, and to the best of our knowledge, no studies have explored their interpretability issues. Addressing this gap can be useful for refining the model's design, in addition to providing new insights about either the model itself (interpretability) or the underlying physical phenomena  (explainability). Models that are inherently interpretable typically have high descriptive accuracy but may sacrifice predictive accuracy on complex datasets compared to black-box models~\cite{linardatos2020explainable}. This trade-off is a key consideration when selecting a model. Black-box models excel when predictive accuracy alone is the primary objective, yet they frequently fall short in scientific contexts where understanding the rationale behind predictions is essential for gaining insights or establishing trust. These models can utilize post-hoc explainability tools that link outputs to inputs meaningfully, including feature attribution, local surrogate, perturbation-based, and gradient-based methods~\cite{wojtas2020feature, ribeiro2016should, lundberg2017unified, sundararajan2017axiomatic, dwivedi2023explainable, zeiler2014visualizing, selvaraju2020grad}. Such methods are popular because they are model-agnostic and do not require modifying the training process. However, recent studies have identified a significant limitation: when applied to the same instance of an opaque model, different post-hoc methods often produce conflicting explanations, highlighting a critical concern regarding their reliability and consistency~\cite{turbe2023evaluation, bello2025level}.


Modeling of complex physical systems demands explainable ML solutions to not only foster trust for ML among scientists, but also to gain new scientific insights. Across various scientific domains, XAI has been increasingly adopted to enhance interpretability in areas such as fluid dynamics, aerospace engineering, neuroscience, molecular science~\cite{cremades2024identifying, sutthithatip2021explainable, menier2025interpretable, proietti2024explainable, sethia2024optimization, zednik2022scientific, li2021kepler}\edit{, as well as in healthcare systems~\cite{wani2024explainable}, network security~\cite{saied2025explainable}, and image processing~\cite{dwivedia2026sfx}.} These efforts highlight the potential of XAI in revealing critical patterns and dependencies within high-dimensional data. However, the choice between by-design (self-interpretable) and post-hoc explainability methods presents trade-offs. While post-hoc methods enable high-accuracy models, the models' reliance on additional tools like SHAP (SHapley Additive exPlanations)~\cite{lundberg2017unified}, and local interpretable model-agnostic explanations (LIME)~\cite{ribeiro2016should} introduces limitations. These tools are computationally intensive for high-dimensional data, lack inherent confidence intervals, and yield explanations that are sensitive to perturbations in the data or model~\cite{jesus2021can, van2022tractability, slack2020fooling}. On the other hand, by-design models, based on pre-defined structures, may limit the discovery of novel patterns and often underperform black-box models in complex tasks due to simplified architecture. Their domain-specific designs restrict generalizability, and they may struggle to capture non-linear interactions, potentially missing subtle scientific phenomena. \edit{Despite the availability of numerous benchmark datasets and increasingly accurate operator-learning architectures, most existing studies continue to evaluate performance primarily in terms of predictive accuracy and generalization, with comparatively little emphasis on intrinsic explainability or interpretable input--output relationship. Although significant progress has been made in identifying important features influencing predictions in scientific modeling, a notable gap remains in developing explainable frameworks for physical systems that are naturally formulated as operator-learning problems. This absence of a robust and self-explanatory framework for modeling functional data and operator-based systems has motivated the development of the present approach.}


\edit{\section{Background and Related Work} \label{sec:Bckgrnd}}

In this work, we build on the integral-equation-based operator-learning methodology introduced in our prior study~\cite{arzani2024interpreting} and extend it to a self-explainable operator-learning approach. In our previous work, the operator learning problem was formulated as a linear combination of generalized functional linear models (gFLM), where the mapping between input and output functions is expressed as a sum of integral equations with learnable kernels and bandwidth parameters. These terms were selected from a predefined library using sparsity-promoting regression, and the resulting model primarily served as an equation-based surrogate for trained neural networks, with a focus on improving out-of-distribution generalization. Here, we build on this work in several important ways. \edit{First, we deliberately remove the nonlinear activation functions applied to the integral terms, allowing the model output to be expressed as an exact linear superposition of contributions. This linear structure is essential for the decomposition-based explainability approach developed here. Second, and most importantly, we introduce a post-training decomposition strategy that partitions the input domain into subregions, computes localized integrals over each region, and explicitly links these input zones to the spatial patterns they generate in the output. This enables simultaneous explainability at both the input and output levels, a capability that was not explored in our earlier work. Finally, we adopt L2 regularization instead of L1 regularization. This choice retains all candidate terms in the model and provides a more stable basis for the mean-centered contribution analysis used in the explainability framework. 
}

\edit{Several features distinguish the proposed framework from existing operator learning and XAI approaches. In DeepONet~\cite{lu2019deeponet}, the operator is factored into a branch network that encodes the input function and a trunk network that encodes the output coordinates. The final prediction is an inner product of the two networks' outputs. While this formulation is mathematically elegant, both components are deep neural networks trained jointly via nonconvex optimization, so the learned basis functions and their coefficients lack closed-form expressions and cannot be traced back to specific regions of the input domain. FNO~\cite{li2020fourier} takes a different route: it applies a Fourier transform to the input at each layer, multiplies selected frequency modes by learnable weight matrices, and transforms back, effectively performing global convolutions in the spectral domain. Although stacking such layers with nonlinear activations between them gives FNO its expressive power, the sequential composition of spectral transforms makes it impossible to isolate how localized input features contribute to the output without external attribution methods. Similarly, kernel-based neural operators~\cite{lowery2024kernel} use kernels with known analytical forms, such as radial basis functions, and approximate the integrals with numerical quadrature. However, the kernel parameters are still optimized through backpropagation within a multilayer neural architecture, so the resulting model inherits the same opacity as other neural operators. In all of these approaches, the operator is learned through nonconvex gradient-based optimization involving thousands to millions of trainable parameters, and the prediction cannot be decomposed into additive, physically localizable contributions. }

In contrast, the proposed gFLM differs on each of these points: the integral terms are defined from a predefined library of analytical kernels and combined in parallel rather than sequentially, and their coefficients are determined through a single convex optimization with a closed-form solution. This linear structure guarantees that the total output can be written exactly as a sum of contributions from distinct input subregions, which directly enables the zonal decomposition without approximation. \edit{It is worth noting that, while functional linear models~\cite{cardot1999functional} based on integral equations have a long history in functional data analysis (FDA)~\cite{horvath2012inference, wang2016functional}, they have traditionally been employed solely as regression tools for prediction. The idea of exploiting the decomposability of the integral operator to provide localized, physically interpretable explanations of the input--output mapping has not, to our knowledge, been explored in prior FDA work. A recent Green's function-based operator network~\cite{gu2025explainable} achieves interpretability by learning the system's Green's function through a DeepONet-inspired architecture and recovering solutions via numerical integration. While this approach provides a physically grounded representation tied to the underlying PDE structure, it still relies on neural networks to approximate the Green's function and its gradients, and is restricted to systems governed by known linear PDEs. Our framework, by contrast, requires no neural network in its pipeline and is applicable to general data-driven mappings between functional data regardless of whether a governing PDE is known. From an XAI perspective, post-hoc methods such as SHAP~\cite{lundberg2017unified} can be applied to neural operators, but they typically provide importance scores only at the input level, require evaluating the model on multiple feature coalitions for each input instance, and operate as an external tool separate from the model's own prediction process. In contrast, explainability in our framework is built into the model itself and extends to the output level by revealing the spatial patterns associated with each input region.} 

Our proposed framework thus contributes to the field of scientific XAI through the following advancements:

\begin{itemize}

\item The integral equation form of our self-explainable operator learning framework provides a closed-form analytical surrogate that not only provides an approximation of functional data but also does so in a self-explainable manner that facilitates flexible explanation of patterns in data, as explained later.

\item The design of our model, a linear combination of integral equations, equips it with inherent decomposability. This intrinsic linearity enables the decomposition of the model's output into distinct components, each corresponding to specific patterns in the input. By quantifying and visualizing these decomposed outputs, our model uniquely links the spatial patterns at the input level to the formation of patterns in the output. This characteristic of our operator-based approach enhances explainability by revealing input-output relationships.  

\item Our framework facilitates the development of XAI metrics, which is an open problem in standardizing AI~\cite{phillips2020four}. By decomposing the output function into mode-like features (similar to modal decomposition techniques), our framework facilitates developing quantitative metrics that could be used to assess the significance of different spatial regions in the input function.

\item Unlike prior XAI models, where explanation occurs on the input (e.g., highlighting important regions in an input image), our framework also naturally reveals the corresponding patterns in the output, providing further explainability. To our knowledge, this is the first XAI approach to deliver explainability at both the input and output levels, offering a novel perspective on output patterns corresponding to different input features.

\item We can apply our XAI framework both by-design (trained from data directly) and post-hoc (trained from a probed neural network). \edit{To validate the explanations produced by our framework, we compare them with those obtained from established post-hoc XAI methods, including Kernel SHAP~\cite{cremades2024identifying}, occlusion sensitivity~\cite{zeiler2014visualizing}, and Gradient-weighted Class Activation Mapping (Grad-CAM)~\cite{selvaraju2020grad}, demonstrating overall agreement across these independent approaches.}

\end{itemize}

\section{Methods} \label{sec:method}


Focusing on physics-based applications, we introduce functional data analysis (FDA) in Sec.~\ref{sec:FDA} as a mathematical basis for mapping in function spaces. Linear kernel-based integral equations were used for formulating an operator in the form of a generalized functional linear model, as detailed in Sec.~\ref{sec:IntpMl}. This FDA-based operator learning framework was developed in our previous work~\cite{arzani2024interpreting}. The novelty of the current study is the post-training analysis, where the decomposed outputs from the model were analyzed to achieve explainability. Here, we omitted the nonlinearity imposed on the integral terms to enable the interpretable decomposition of the output of the integral operator. Therefore, the proposed framework here trades some expressivity for explainability, as described in Sec.~\ref{sec:Physics}. Three post-hoc explainability methods were introduced in Sec.~\ref{sec:comp} as baseline methods for comparison with the proposed framework. Finally, two test cases were selected in  Sec.~\ref{sec:tests} to assess the model’s performance and explainability in physics-based contexts.

\edit{
\noindent The overall workflow of the proposed framework is summarized in Figs.~\ref{fig:Schematic} and~\ref{fig:Corr}. The methodology consists of two stages: (i) training of the XAI model (Fig.~\ref{fig:Schematic}) and (ii) post-training inference and decomposition analysis (Fig.~\ref{fig:Corr}), summarized below. 

\begin{itemize}
\item In the first stage, functional input--output datasets are collected from the physical system of interest and represented on discretized spatial domains. A library of candidate integral operators is then constructed using multiple kernel functions, bandwidth parameters, and lifted transformations of the input functions. The resulting candidate integrals are assembled into a feature library matrix, and the coefficients of the generalized functional linear model (integral operator terms) are determined through regularized linear regression.

\item In the second stage, the trained integral operator is used to analyze individual samples of interest. The input domain is partitioned into four equal subregions, and the feature library is recomputed separately for each zone using only the feature values and spatial coordinates contained within that subdomain. Using the learned coefficients of the integral operator, localized predictions associated with each zone are obtained independently. Finally, a mean-centering procedure is applied to these zonal predictions to compute the zonal contributions and reveal how different spatial regions of the input influence the final output prediction.
\end{itemize}
}

\subsection{Functional Data Analysis (FDA)} \label{sec:FDA}

FDA treats observed data as continuous functions rather than discrete measurements~\cite{wang2016functional}. In many scientific problems, data take the form of smooth functions over a continuous domain, suggesting the potential for specialized regression techniques. Functional regression aims to learn relationships in which at least one of the predictor or response variables is functional. Neural operators extend this idea using highly expressive neural-network architectures~\cite{kovachki2023neural}, but these models are typically opaque and difficult to interpret. In contrast, functional linear models (FLMs) provide a more interpretable formulation based on integral equations~\cite{wang2016functional, horvath2012inference}. In this study, we consider two FLM formulations: function-to-function mapping and function-to-scalar mapping~\cite{arzani2024interpreting}.

\subsubsection{Fully Functional Model (Function-to-Function)}

In the fully functional model, both the input $\mathbf{f}(\bm{\xi})$ and the output $\mathbf{u}(\bx)$ are functions, that is, they vary continuously over their domains. The regression relationship is given by:

\begin{equation} 
\mathbf{u}(\bx) = \int  \bm{\psi}(\bx, \bm{\xi}) \mathbf{f}(\bm{\xi})   \,d \bm{\xi},
\label{eqn:fda1}
\end{equation}
where $\bm{\psi}(\bx, \bm{\xi})$ is a spatially dependent kernel function (or a collection of kernels) that determines how different points in $\mathbf{f}(\bm{\xi})$ contribute to $\mathbf{u}(\bx)$ at each spatial location $\mathbf{x}$.

\subsubsection{Scalar Response Model (Function-to-Scalar)}

In the scalar response model, the input remains a function, but the response is a functional in the form of a single scalar value. The regression equation in this case is represented as:

\begin{equation} 
\mathbf{u} = \int  \bm{\psi}(\bm{\xi}) \mathbf{f}(\bm{\xi})   \,d \bm{\xi}, 
\label{eqn:fda2}
\end{equation}
where the kernel $\bm{\psi}(\bm{\xi})$ is a weight function that determines the contribution of different parts of $\mathbf{f}(\bm{\xi})$ to the scalar output. This model can be viewed as a functional generalization of multiple linear regression, where the sum of weighted features is replaced by an integral operator~\cite{arzani2024interpreting, mclean2014functional}.

\subsubsection{Lifting Maps}

To improve the expressive capability of FLMs, input functions can be transformed into higher-dimensional feature spaces through lifting transformations~\cite{qian2020lift}. A lifting transformation $\mathcal{T}$ is defined as:

\begin{equation} 
\mathcal{T}: \mathbf{f}(\bm{\xi}) \mapsto \mathcal{T} \mathbf{f}(\bm{\xi}),
\label{eqn:transformation}
\end{equation}
where $\mathcal{T}$ denotes a pre-specified nonlinear transformation. Examples include polynomial, exponential, and hyperbolic tangent mappings~\cite{gosea2022exact, viknesh2026adam}. Applying these transformations allows nonlinear dependencies in the original data to be represented through an expanded feature space, while the model could still be kept linear by considering these lifted functions as new features. The lifted functions $\mathcal{T}\mathbf{f}(\bm{\xi})$ are incorporated into the operator as:

\begin{equation} 
\mathbf{u}(\bx) = \sum_{n=1}^{N}\sum_{m=1}^{M} \int  \bm{\psi}_n(\bx, \bm{\xi}) \mathcal{T}_m\mathbf{f}(\bm{\xi})   \,d \bm{\xi}, 
\label{eqn:fda1andT}
\end{equation}
where $N$ different kernels are discovered for $M$ lifted function. This approach aligns with the Koopman operator framework and kernel-based operator learning, where nonlinear systems are embedded into a feature space where their dynamics appear linear~\cite{mezic2013analysis, williams2015kernel}.

\subsubsection{Kernel Functions}

A library of various kernel functions was considered in this work for the operator learning task.  The form of these kernel functions was previously expressed in our prior work~\cite{arzani2024interpreting}, and a summary is presented in the Appendix (Sec.~\ref{sec:kernels}).

\subsection{Generalized Functional Linear Model (gFLM)} \label{sec:IntpMl}

Model selection and hyperparameter tuning remain challenging in kernel-based regression models because the choice of kernel function and bandwidth strongly influences the learned functional relationships~\cite{kohler2014review, altman1995bandwidth}. The difficulty arises because different kernels capture different structures in data, and there is no universal selection criterion that guarantees optimal performance across diverse datasets. Instead of pre-specifying a single kernel function $\bm{\psi}(\bx, \bm{\xi})$, we adopt a library-based approach consisting of multiple candidate kernels and bandwidth parameters~\cite{arzani2024interpreting}. Most kernels in the library include a bandwidth parameter controlling the locality and smoothing properties of the operator. To avoid relying on a fixed problem-dependent bandwidth, multiple candidate bandwidths are considered for each kernel function.

Finally, by incorporating $N$ kernels and $M$ lifting maps, we solve the operator learning problem using integral equations by integrating information from nearby points in a structured manner. The resulting framework can represent both local and nonlocal dependencies within the functional data. The general forms of the resulting function-to-function and function-to-scalar models for 2D domains are given by:

\begin{equation} 
\mathbf{u}({x}, {y}) = \sum_{n=1}^{N} \sum_{m=1}^{M} w_{n,m}\iint \bm{\psi}_n({x}-{\zeta}, {y}-{\eta}) \mathcal{T}_m \mathbf{f}({\zeta},{\eta})   \,d \zeta d \eta, \; \;  \textup{function-to-function mapping}
\label{eqn:fda1T2D}
\end{equation}

\begin{equation} 
\mathbf{u} = \sum_{n=1}^{N} \sum_{m=1}^{M} w_{n,m} \iint  \bm{\psi}_n({\zeta}, {\eta}) \mathcal{T}_m \mathbf{f}({\zeta},{\eta})   \,d \zeta d \eta, \; \;  \textup{function-to-scalar mapping} 
\label{eqn:fda2T2D}
\end{equation}
where $w_{n,m}$ denotes the learned coefficient associated with each integral term. Although the framework can be extended to vector-valued fields, the present study focuses on scalar fields.


After defining the candidate kernels and associated hyperparameters, the operator-learning problem is formulated as a regression problem over the library of integral terms. No constraints are imposed on the kernel functions other than integrability, allowing flexibility in the kernel selection. The resulting gFLM represents the proposed explainable AI (XAI) framework and is expressed as a linear superposition of integral operators. This gFLM is considered mathematically interpretable because it employs a surrogate model in the form of a transparent linear integral operator, where the output is directly computed by weighting the input function with kernel functions~\cite{arzani2024interpreting}. The framework is explainable since the integral equation nature of the model could be utilized to linearly decompose the contributions from different regions of the input function and demonstrate how different regions in the input are mapped to the patterns observed in the output. We argue that our model is ``self-explainable'' because, unlike other explainability models (e.g., SHAP (SHapley Additive exPlanations)), it does not require additional post-processing steps, and the model itself naturally leads to the explanation, which will be discussed later in Section~\ref{sec:Physics}.

\subsubsection{Functional Regression Formulation} 

To determine the unknown coefficients associated with each integral term in the defined library of candidate integral equations, we formulate a linear system of equations based on the integral operator models~\cite{arzani2024interpreting}. Given pairs of training data and considering the gFLM model in the previous section, we define the training set in vector form as:

\begin{equation}
\mathbf u(\bx) = [u_1(\mathbf{x}), u_2(\mathbf{x}), \ldots, u_q(\mathbf{x})]^T,
\end{equation}

\begin{equation}
\mathbf f(\bm{\xi}) = [ f_1(\bm{\xi}), f_2(\bm{\xi}), \ldots,  f_q(\bm{\xi})]^T,
\end{equation}

\begin{figure}[h!]
\centering
\includegraphics[scale=0.290]{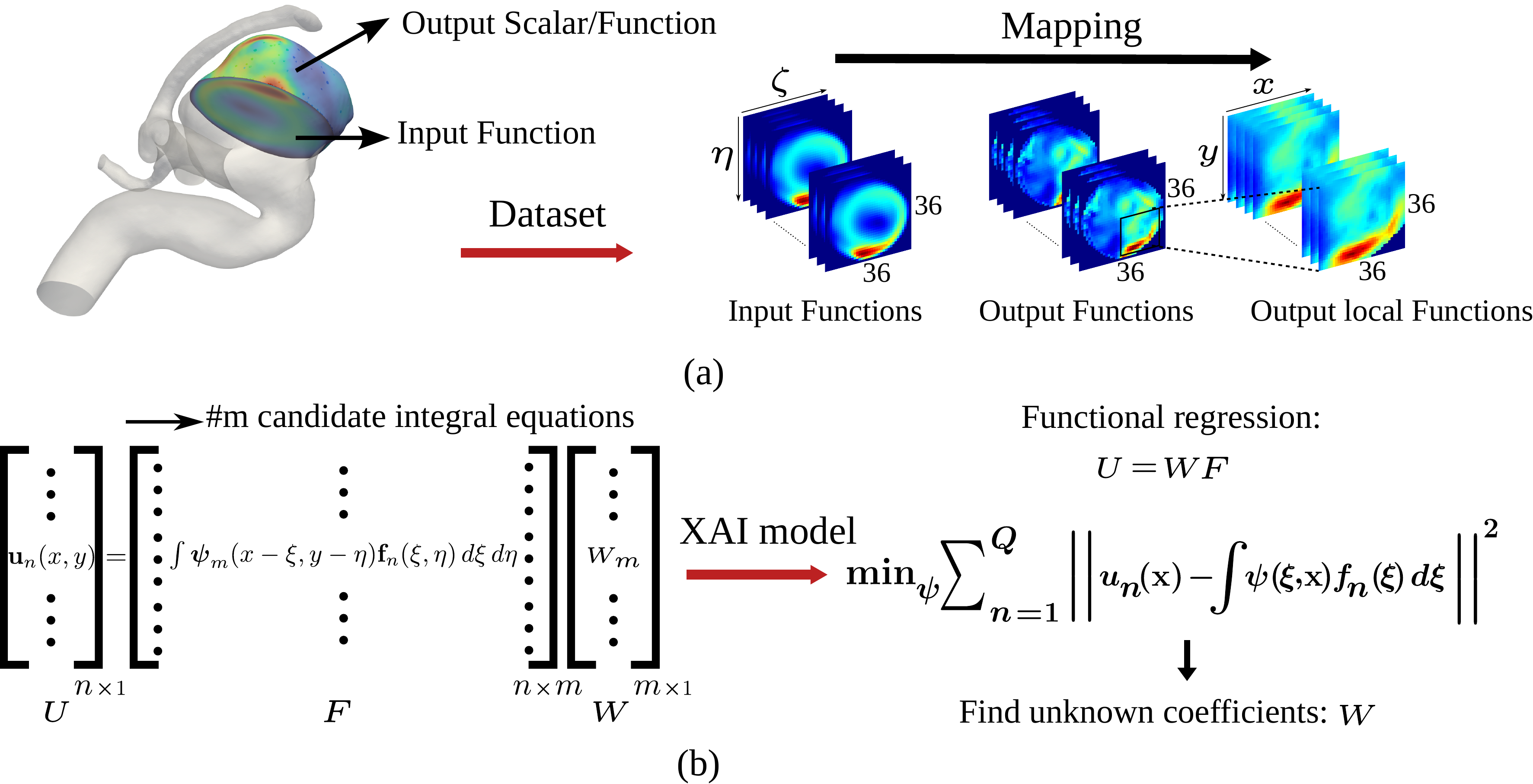}
\caption{XAI model training process. (a) Blood flow inside a brain aneurysm is chosen for functional mapping. The velocity magnitude field on a 2D plane in the aneurysm area is used as the input function to the model. The output is wall shear stress (WSS)  magnitude extracted from the corresponding projected wall with respect to the 2D plane. Specifically, the localized high-magnitude WSS field with the same resolution is selected as the output function of the model. (b) A library of candidate integral equations is constructed to conduct the mapping from the input candidate integral equations and function ($F$) to the output scalar/function ($U$). This integral operator is then probed by using the dataset and formulating a convex optimization problem to solve for the integral terms' coefficients.}
\label{fig:Schematic}
\end{figure}

As shown in Fig.~\ref{fig:Schematic}, an input function $\mathbf f(\bm{\xi})$ is mapped to either another function, $\mathbf u(\bx)$, or a scalar value, $\mathbf{u}$. The integration over the input function domain could be performed using any numerical method for any candidate term in the library, resulting in a known matrix $\mathbf{F}$ representing the library of integral equations (Fig.~\ref{fig:Schematic}). Therefore, a system of linear equations is formulated as:

\begin{equation}
\mathbf{U}=\mathbf{F}\mathbf{W},
\end{equation}
where $\mathbf{U}$ contains the output functions in the training data and $\mathbf{W}$ contains the unknown coefficients of the candidate integral terms. To solve for $\mathbf{W}$, we define a convex optimization problem:

\begin{equation}
\min_{\mathbf{W}} \, \| \mathbf{U} - \mathbf{F}\mathbf{W} \|_2,
 \label{eqn:eq1}
\end{equation} 
The normal equations provide a closed-form analytical solution to the unregularized least-squares regression problem:

\begin{equation} 
\mathbf{F}^T \mathbf{F} \mathbf{W}  =  \mathbf{F}^T \mathbf{U},
\end{equation}

\noindent Unlike our previous work~\cite{arzani2024interpreting}, which encouraged sparsity on $\mathbf{W}$ by applying L1 norm, here we utilize the L2 norm to make the training process more stable:

\begin{equation}
\min_{\mathbf{W}} \, \| \mathbf{U} - \mathbf{F}\mathbf{W} \|_2 + \lambda \| \mathbf{W} \|_2,
 \label{eqn:eq2}
\end{equation}  

\begin{equation} 
( \mathbf{F}^T \mathbf{F} + \lambda \mathbf{I} ) \mathbf{W}  =  \mathbf{F}^T \mathbf{U},
\end{equation}
where $\lambda$ is the regularization parameter. This method penalizes large coefficients and keeps them small, which is helpful when calculating the outputs for the decomposed integral equations. The L1 norm choice will lead to a mathematically sparse and more interpretable model, whereas for the purpose of explaining the input-to-output mapping patterns, the sparse nature of the integral equation is not a concern. As shown next, the self-explainability only relies on the integral equation nature of the model and not the number of terms.




\subsection{Self-explainable Operator} \label{sec:Physics}

The surrogate XAI model, resulting from defining an operator learning problem, provides a closed-form solution that maps any input function $\mathbf{f}$ to an output function $\mathbf{u}$. The XAI model is expressed as a sum of integral equations, each weighted by coefficients that are obtained as explained in the previous section. Here, we demonstrate the self-explanatory nature of the model. For simplicity, below, we will consider one generic term in Eq.~\ref{eqn:fda1T2D} and \ref{eqn:fda2T2D}, and the addition of other terms based on different kernels and lifted input functions will simply have an additive effect. Additionally, we present the function-to-function mapping, and the function-to-scalar mapping follows the same logic.

To understand the spatial relationships between the input and output data, we analyze the integral operator by decomposing the input domain into $k$ different subdomains $\Omega_T = \bigcup_{i=1}^{k} \Omega_i$. This allows us to express the total integral evaluation as a sum of localized integrals, computed over each subdomain:

\begin{equation} 
\mathbf{u}_i({x}, {y}) = \iint_{\Omega_i} \bm{\psi}({x}-{\zeta}, {y}-{\eta}) \mathcal{T}\mathbf{f}({\zeta},{\eta})   \,d \zeta d \eta, 
\end{equation}
where $\Omega_i$ represents different subdomains in the input space and  $\mathbf{u}_i({x}, {y})$ is the contribution to the output coming from the $\Omega_i$ region. By integrating over an arbitrary subdomain, we obtain a localized contribution to the final XAI model prediction that corresponds to that specific region.

\noindent Due to the linear property of integral equations, we can express the final XAI model prediction as the sum of contributions from different subdomains:

\begin{equation} 
\mathbf{u}({x}, {y}) =  \iint_{\Omega_T} \bm{\psi}({x}-{\zeta}, {y}-{\eta}) \mathcal{T}\mathbf{f}({\zeta},{\eta})   \,d \zeta d \eta  \ = \sum_{i=1}^{k} 
 \iint_{\Omega_i} \bm{\psi}({x}-{\zeta}, {y}-{\eta}) \mathcal{T}\mathbf{f}({\zeta},{\eta})   \,d \zeta d \eta  = \sum_{i=1}^{k} \mathbf{u}_i({x}, {y}),
\label{eqn:eq15}
\end{equation} 

This equation demonstrates the self-explanatory nature of our XAI model. Namely, due to the mathematical nature of the model, the output could be decomposed into a simple sum of $k$ contributions coming from $k$ different, arbitrarily selected subregions of the input as long as the subregions cover the entire input domain. In this study, $k=4$ is used to divide the input domain function into four equal zones. Each term in this sum (Eq.~\ref{eqn:eq15}) represents the prediction of the XAI model of a specific zone, each of which contributes separately to the overall XAI model prediction. By plotting each contribution $\mathbf{u}_i$ coming from zone $\Omega_i$ and keeping in mind the linear nature of the final model, this approach provides a physically interpretable way to analyze how different regions of the input domain influence the formation of patterns in the output function.

\edit{The choice of $k=4$ represents a practical compromise between spatial localization and interpretability. A sensitivity analysis presented in Appendix (Sec.~\ref{sec:app_zones}) examines the effect of varying $k$ on the decomposition results using $k=2$, $k=4$, and $k=9$. Across all partitions, the same dominant spatial pattern emerges in the output contributions: with $k=2$, it appears in one of the two zones depending on the partition orientation; with $k=4$, it is cleanly isolated in a single zone; and with $k=9$, it is distributed across adjacent zones that overlap with the same physical region. Using $k=4$ provides sufficient resolution to localize the dominant input features while keeping the number of zones small enough for clear interpretation.}

To better assess how each zone in the input function impacts the predictions made by the XAI model in the mapping, we employ the following formulation:

\begin{equation}
\Phi_i = \mathbf{u}_i({x}, {y}) - E[\mathbf{u}_i({x}, {y})] \;,
\end{equation}
where the mean-centered contribution $\Phi_i$ of each zone's output to the prediction is calculated as the difference between the local XAI prediction derived from integrating over that zone and the expected value of XAI prediction computed for the same zone across all dataset instances. This mean-centering strategy significantly enhances the comparison of the zonal contributions. An analogy could be made with principal component analysis (PCA), where the data is mean-centered before singular value decomposition (SVD) is applied, and the modes derived are mean-centered. Additionally, a similar mean-centering approach is used in the Shapley value theory~\cite{molnar2020interpretable}. It is simple to show that once the mean-centered contributions are summed, the mean-centered output (the total output function minus its average) is recovered:

\begin{equation}
\sum_{i=1}^{k} \Phi_i = \mathbf{u}({x}, {y}) - E[\mathbf{u}({x}, {y})].
\end{equation}

\begin{figure}[h!]
\centering
\includegraphics[scale=.32]{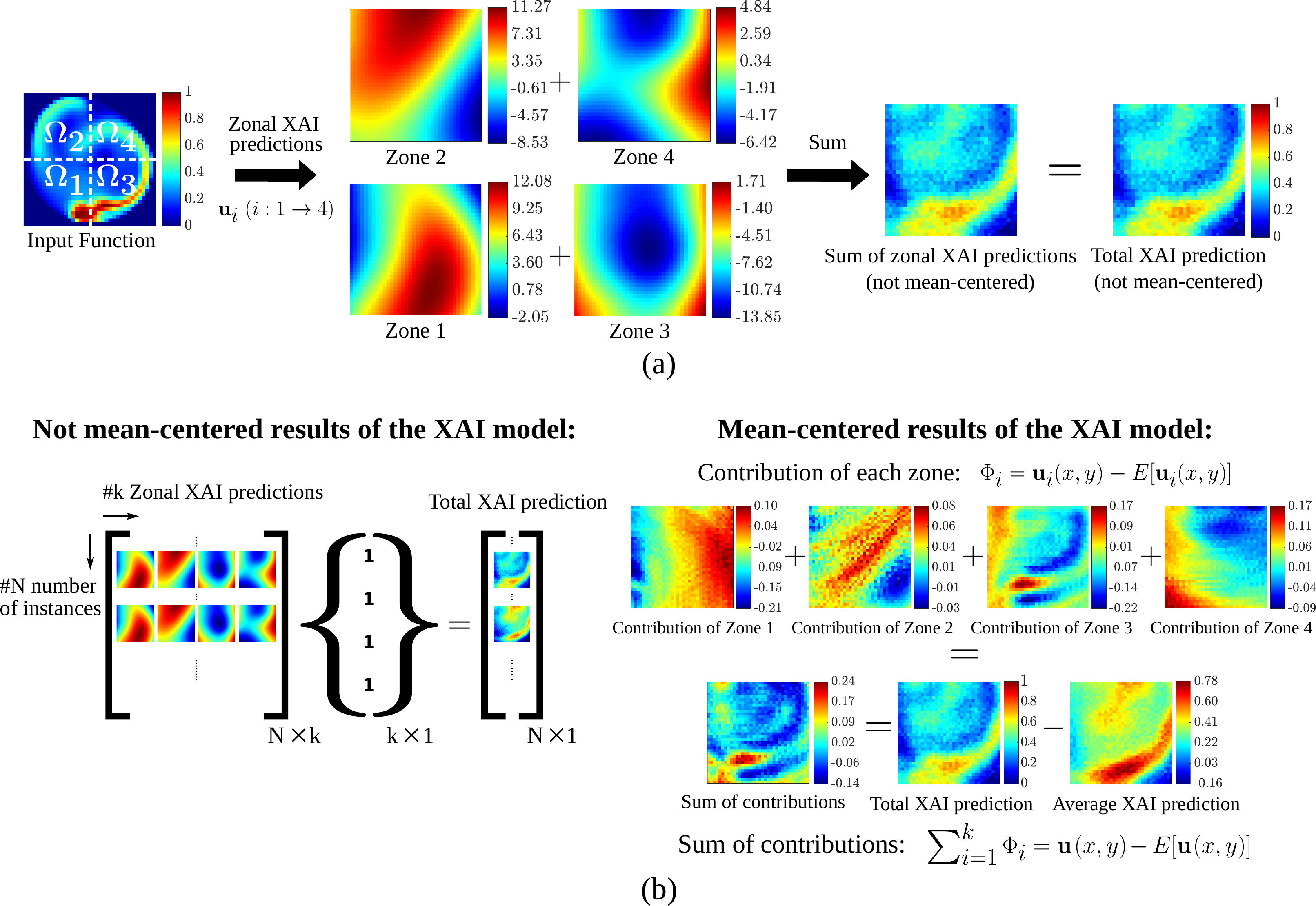}
\caption{\edit{Post-processing procedure for the proposed XAI framework. (a) The input function domain for a representative sample from the blood flow function-to-function mapping task is partitioned into four equal subdomains, $\Omega_1$--$\Omega_4$. Using the trained integral operator, the XAI model independently evaluates each subdomain to produce localized predictions (not mean-centered). Summing these four predictions recovers the full XAI model prediction obtained by integrating over the entire input domain. (b) Mean-centered zonal contributions are computed by subtracting the expected value of each zonal prediction, evaluated across all dataset instances, from its corresponding prediction at the selected sample. Owing to the additive structure of the integral operator, the sum of the mean-centered zonal contributions exactly recovers the mean-centered XAI model prediction.}}
\label{fig:Corr}
\end{figure}

\edit{It should be noted that the expectation operator in the mean-centering procedure is computed over a predefined collection of input functions selected for interpretation. In the present study, this set corresponds to the test dataset used for evaluating the trained XAI model.} For each input instance in this set, the corresponding zonal prediction $\mathbf{u}_i({x},{y})$ is first computed using the trained operator. The expectation $E[\mathbf{u}_i({x},{y})]$ is then evaluated (separately for each zone) as the sample mean across all considered instances and subtracted from the zonal prediction to obtain the mean-centered zonal contribution. \edit{Consequently, the resulting mean-centered contributions $\Phi_i$ depend on the distribution and characteristics of the selected dataset used for interpretation.} The procedure conducted to compute the zonal contributions from the XAI model output for each of the four zones is illustrated in Fig.~\ref{fig:Corr} for a representative snapshot.

To quantitatively assess the zonal contribution patterns represented by $\Phi_i$, the Pearson correlation coefficient (PC) is computed between the contribution from each zone and the overall XAI model prediction. \edit{For function-to-scalar mappings, this yields a single PC value for each zone across the dataset.} In the function-to-function mapping case, the correlation is computed separately for each sample, and the resulting PC values are subsequently averaged over all input instances to obtain a single quantitative metric describing the relative influence of each zone on the predicted output. This metric quantifies the significance of existing spatial patterns in the input domain function in shaping the output patterns. 

A beneficial characteristic of integral operators is that the integration over the full domain can be decomposed into smaller sub-integrations over distinct regions. The sum of these regional integrations equals the integral over the entire input domain. Exploiting this property, we first train the XAI model (integral operator learning) using the dataset and subsequently divide the input domain into four distinct subdomains or zones. Using the learned integral operator, we separately integrate the input function over each of these four zones to obtain individual contributions from each region. By computing these localized contributions, we quantify and visualize how information in each zone influences the formation of the final output. Since each zone exhibits a distinct spatial pattern, this approach enables us to determine which patterns or (physical) features in the input domain are most crucial for yielding the output. Ultimately, this methodology provides explainability at both the input and output levels, since each regional integral directly represents a portion of the final predicted output.

\subsection{Comparison with Post-Hoc Explainability Methods} \label{sec:comp}

To benchmark the explanations produced by the proposed framework against established post-hoc XAI approaches, we trained a U-Net model~\cite{ronneberger2015u} on the same function-to-function mapping task considered in the blood flow test case. The network maps a single-channel $36\times36$ input function to a corresponding $36\times36$ output function through an encoder--decoder architecture with skip connections. The encoder consists of four downsampling stages with convolutional blocks and max-pooling operations, progressively increasing the number of feature channels from \edit{64 to 512,} while the decoder symmetrically upsamples the latent representation using bilinear interpolation and convolutional layers. The final output is generated through a $1\times1$ convolution layer. The model was trained with mean squared error (MSE) loss, the Adam optimizer (learning rate $=10^{-3}$, weight decay $=10^{-5}$), and a ReduceLROnPlateau scheduler with factor 0.5 and \edit{patience 50.} Early stopping was applied with a patience of 200 epochs based on the test loss, and training was run for at most 5000 epochs with a batch size of \edit{32.} After training, the checkpoint with the lowest test loss was retained for the explainability analysis.

\subsubsection{Kernel SHapley Additive exPlanations (Kernel SHAP)} 

The U-Net architecture, similar to other opaque neural networks, requires an additional explainability model to reveal its prediction mechanisms and the relationships between inputs and outputs (unlike our self-explainable XAI model). A post-hoc model uses the parameters of the trained U-Net to analyze both the inputs and the resulting predictions. Additive feature-attribution techniques model the system as a linear framework, where the discrepancy between the U-Net's output and the ground truth data is approximated by a simpler explanation model, denoted as \edit{$g(z)$}:

\edit{
\begin{equation}
g(z) = \Theta_0 + \sum_{i=1}^{N} \Theta_i z_i,
\end{equation}}
where $\Theta_0$ represents the model's output when all features are excluded, $\Theta_i$ denotes the SHAP values associated with each feature, and \edit{$z_i \in \{0,1\}$} indicates whether feature $i$ is included in the coalition. The vector \edit{$z$} therefore represents a binary coalition of active and inactive features.

The Kernel SHAP algorithm~\cite{cremades2024identifying} evaluates the model’s behavior by sampling subsets of features (coalitions) and observing how the model’s output changes when certain features are included or excluded. SHAP values are calculated by minimizing a loss function $\mathcal{L}$, which depends on $f(h(z'))$, the model’s output for a given feature coalition $z'$, $g(z')$, the explanation model’s output, and $\pi_x(z')$, the kernel weight, which prioritizes coalitions based on their size. The space of the input functions is converted to a space of binary values using a mask function $h$:

\begin{equation}
\mathcal{L} = \sum_{z' \in \mathcal{Z}} \left( f(h(z')) - g(z') \right)^2 \pi_x(z'),
\end{equation}
and the kernel $\pi_x(z')$ is defined as:

\begin{equation}
\pi_x(z') = \frac{N-1}{\binom{N}{|z'|} |z'| (N - |z'|)},
\end{equation}
where $N$ is the total number of features and $|z'|$ is the number of features included in the coalition. To ensure a consistent comparison between the predictions of our surrogate XAI model and Kernel SHAP, we divided the input function domain into four equal zones and treated each as an individual feature influencing the predictions to derive corresponding SHAP values, where the absent zone of each coalition is replaced by a background of zero values. The final SHAP value for each zone is computed by averaging the importance scores from all coalitions in which it was included. The importance is calculated using the squared error between the model’s prediction and the ground truth. The most significant zone, according to SHAP, is the one exhibiting higher absolute values, whereas zones with lower absolute values are considered less relevant. 

\edit{
\subsubsection{Occlusion Sensitivity} 

To further benchmark the explanations produced by the proposed framework, we applied an occlusion sensitivity analysis~\cite{zeiler2014visualizing} to the trained U-Net model. Occlusion sensitivity is a perturbation-based explainability method that evaluates the importance of an input region by measuring how the model prediction changes when that region is removed or masked.

To maintain consistency with the zonal decomposition used throughout this work, the input domain was partitioned into the same four equal subregions employed in the proposed XAI framework. For each test sample, the original prediction error was first computed using the MSE between the predicted output and the ground-truth function. One zone was occluded at a time by replacing the corresponding region in the input velocity field with zero values, while the remaining regions were left unchanged. The occluded input was then passed through the trained U-Net model, and the prediction error was recomputed.

The importance of each zone was quantified by the increase in MSE relative to the original unoccluded prediction:

\begin{equation}
\Delta_{zone_i}
=
\mathrm{MSE}_{\mathrm{occluded},i}
-
\mathrm{MSE}_{\mathrm{original}},
\end{equation}
where $\mathrm{MSE}_{\mathrm{original}}$ is the prediction error obtained using the complete input field and $\mathrm{MSE}_{\mathrm{occluded},i}$ is the error after masking zone $i$. Larger values of $\Delta_{zone_i}$ indicate that the model relies more strongly on that region for prediction, since removing the region causes greater degradation in performance. The final occlusion scores were computed independently for all test samples and compared with the zonal ranking obtained from the proposed framework.
}

\edit{
\subsubsection{Gradient-weighted Class Activation Mapping (Grad-CAM)} 
Grad-CAM~\cite{selvaraju2020grad} was also applied to the trained U-Net model as an additional post-hoc explainability method. Unlike Kernel SHAP and occlusion sensitivity, which estimate feature importance through perturbation or coalition sampling, Grad-CAM generates saliency maps directly from the gradients of the neural network. Specifically, the method computes the gradients of the prediction with respect to feature maps in a selected convolutional layer and uses them to generate spatial activation maps highlighting the regions that contribute most strongly to the output. In this study, Grad-CAM was evaluated using multiple encoder stages of the U-Net in order to investigate how the spatial granularity of the explanations changes with network depth. Additional implementation details and Grad-CAM results are provided in Appendix (Sec.~\ref{sec:gradcam}).
}

\subsection{Test Problems} \label{sec:tests}

We employ the presented self-explainable operator to interpret how the operator maps physical variables. Two distinct tasks are defined to achieve this: function-to-scalar mapping and function-to-function mapping. The test problems presented below include pulsatile blood flow in cardiovascular disease and unsteady aerodynamics. In both cases, training is done using a spatiotemporal dataset based on a given simulation, where the snapshots are considered the training data, and the goal is to map spatial patterns as explained below. In practice, this corresponds to a local interpretation of a model that might be trained across different boundary conditions and parameters, and one is interested in explaining the results for a given scenario (fixed boundary condition and parameters).

\subsubsection{Blood Flow Dataset} 

The dataset was derived from a blood flow simulation within a patient-specific internal carotid artery (ICA) brain aneurysm in a prior study~\cite{csala2022comparing}. SimVascular was used for mesh generation and simulation, with a computational mesh of 6.6 million elements, including a boundary layer mesh. A pulsatile inflow waveform was applied at the inlet, split-resistance outlet boundary conditions were imposed based on Murray's law, and the vessel walls were modeled as rigid with a no-slip condition. Blood was treated as a Newtonian fluid ($\rho=1060\,\text{kg/m}^3$, $\mu=0.004\,\text{kg/ms}$, $Re=322$). Three cardiac cycles were simulated, with 1,001 snapshots extracted from the final cycle, cropped, and downsampled to a coarser mesh of 40,500 points. \edit{Given the importance of wall shear stress (WSS) in cardiovascular fluid mechanics~\cite{Mahmoudi2021story}, we focused on predicting WSS quantities as the output.} Utilizing velocity magnitude fields from a planar slice in the aneurysm region as input functions, we performed two mapping tasks: function-to-scalar and function-to-function. For function-to-scalar mapping, the XAI model was trained separately on three output quantities extracted from the projected wall: maximum WSS (Max WSS), spatially averaged WSS (Mean WSS), and the spatial average of divergence of normalized WSS (Mean DNWSS)~\cite{morbiducci2020wall}. For function-to-function mapping, the output is the localized high-WSS field defined over the coordinate intervals $0.45<x<0.9$ and $0.05<y<0.5$ (Fig.~\ref{fig:Schematic}). The spatial input domain coordinates are represented by $0<\zeta<1$ and $0<\eta<1$. \edit{Further details regarding the preprocessing and interpolation for the blood flow dataset are provided in Appendix (Sec.~\ref{sec:preprocessing}).}

\edit{For the function-to-scalar tasks, 500 evenly spaced snapshots were selected from the 1,001 available time steps. For the function-to-function task, 300 evenly spaced snapshots were selected from the full set. Input functions were normalized per sample to $[0, 1]$ using the minimum and maximum values of each individual sample. For function-to-scalar mappings, the scalar output quantities were normalized globally across the dataset, whereas for function-to-function mappings, the output functions were also normalized per sample. To mitigate temporal correlation among consecutive snapshots, 400 and 200 samples were randomly drawn for training in the function-to-scalar and function-to-function tasks, respectively, with 100 samples reserved for testing in each case. This random splitting was repeated across five realizations to assess uncertainty, and all results are reported as mean $\pm$ standard deviation.} A diverse library of candidate integral terms was constructed using exponential-type kernel functions with a broad range of bandwidth hyperparameters ($\beta$), and the regularized linear regression was solved via the GMRES iterative solver with an L2 regularization parameter $\lambda$. Hyperparameter selections are summarized in Table~\ref{tab:Hyp}.

\subsubsection{Airfoil Dataset} 

Data-driven methods have been proven to be accurate and fast for predicting the drag coefficient~\cite{shi2025drag}. As a second test case, we used the unsteady aerodynamics database of Towne et al.~\cite{towne2023database}, who conducted direct numerical simulations for a two-dimensional flat plate airfoil at a Reynolds number of $100$. A sinusoidal pitch was imposed about a base angle of attack ($\alpha_0 = 25^\circ$) following:
\begin{equation}
\alpha(t) = \alpha_0 - \alpha_p \sin(2 \pi f_p t),
\end{equation}
where $\alpha_p = 5^\circ$ is the pitching amplitude and \edit{$f_p = f_c / U_\infty$} is the dimensionless pitching frequency, ranging from \edit{0.05 to 0.5;} $f_c$ is the pitching frequency in Hz and $U_\infty$ is the freestream velocity. The simulations employed an immersed boundary projection method for incompressible flow, with Crank-Nicolson time integration for linear terms and a third-order Runge-Kutta scheme for nonlinear terms. Velocity fields were recorded on a uniform spatial grid with spacing $0.02c$ ($c$ is the chord length) at a temporal resolution of $0.1c/U_\infty$, yielding 401 snapshots for the $\alpha_0 = 25^\circ$ case. Time-varying drag ($C_d$) and lift ($C_l$) coefficients were also collected at a finer temporal resolution of $0.01c/U_\infty$.

For the function-to-scalar mapping task using this dataset, we considered two dimensionless pitching frequencies, $f_p = 0.05$ and $f_p = 0.2$. All 401 velocity snapshots were downsampled to $100 \times 100$ spatial resolution through interpolation and used as input functions to the XAI model, with the corresponding $C_d$ and $C_l$ values as scalar outputs. \edit{The spatial coordinates of the input domain are denoted by $\zeta$ and $\eta$, with both coordinates defined over the interval $[0, 1]$. Following the same protocol as the blood flow test case, the data were randomly split into training (80\%) and test (20\%) sets across five random seeds, and all results are reported as mean $\pm$ standard deviation.} The XAI model hyperparameters are summarized in Table~\ref{tab:Hyp}.

\edit{Additional implementation details regarding numerical integration, computational complexity, and runtime performance are provided in Appendix (Sec.~\ref{sec:complexity}). Moreover, a sensitivity analysis examining the robustness of the model predictions and zonal explanations to hyperparameter choices (kernel bandwidth ($\beta$), regularization strength ($\lambda$)) and input noise perturbations is provided in Appendix (Sec.~\ref{sec:sensitivity}).}

\renewcommand{\arraystretch}{1.2}

\begin{table}[h]
    \caption{Hyperparameter tuning in the mapping tasks. Here, $\beta$ is the bandwidth hyperparameter in exponential-type kernel functions. In the library of candidate terms, each kernel is treated independently, and for each such kernel, multiple candidate bandwidths are considered. Specifically, the range shown is uniformly discretized into $P$ values.
    $\lambda$ is the L2 regularization parameter in Eq.~\ref{eqn:eq2}.
    For function-to-scalar mappings, a spatial localization parameter $h$ is introduced to shift the kernel center across the input domain and capture localized contributions of different regions. The parameter $h$ is varied over the interval $[0, 1]$, discretized with the step size listed.}
    \label{tab:Hyp}
    \centering
    \small
    \setlength{\tabcolsep}{4pt}
    \renewcommand{\arraystretch}{1.1}
    
    \begin{tabular}{llccc}
    \toprule
    \textbf{Test case} & \textbf{Task} & \bm{$\beta$} & \bm{$\lambda$} & \textbf{step size} (\bm{$h$}) \\
    \midrule
    \multirow{4}{*}{Blood flow} 
        & Max WSS                  & $[0.1, 2.5],\,P=100$ & $10^{-5}$ & 0.2 \\
        & Mean WSS                 & $[0.1, 2.5],\,P=100$ & $10^{-5}$ & 0.2 \\
        & Mean DNWSS               & $[0.1, 2.5],\,P=120$ & $10^{-5}$ & 0.2 \\
        & Function-to-Function     & $[0.2, 1.5],\,P=120$ & $10^{-9}$ & --  \\
    \midrule
    \multirow{4}{*}{Airfoil} 
        & $f_p=0.05$, $C_d$        & $[0.1, 2.5],\,P=100$ & $10^{-5}$ & 0.2 \\
        & $f_p=0.05$, $C_l$        & $[0.1, 2.5],\,P=100$ & $10^{-5}$ & 0.2 \\
        & $f_p=0.2$, $C_d$         & $[2, 4],\,P=50$      & $10^{-5}$ & 0.3 \\
        & $f_p=0.2$, $C_l$         & $[2, 4],\,P=50$      & $10^{-5}$ & 0.3 \\
    \bottomrule
    \end{tabular}
    
\end{table}

\edit{\subsection{Fourier Neural Operator Baseline} \label{sec:fno_baseline}}

\edit{To quantitatively examine the accuracy--interpretability trade-off discussed earlier, we trained a Fourier Neural Operator (FNO)~\cite{li2020fourier} as a baseline on all mapping tasks. Additional implementation details regarding the FNO architectures, training procedures, and hyperparameter configurations are provided in Appendix (Sec.~\ref{sec:fno}).}

\section{Results} \label{sec:results}



In the following, we evaluate the proposed XAI framework using both quantitative and qualitative analyses on unseen test data from the two problem sets. The zonal contributions are ranked using Pearson correlation coefficients computed between the mean-centered output associated with each input region and the overall XAI model prediction, allowing the dominant spatial patterns influencing the output to be identified. \edit{A quantitative comparison with the FNO baseline is also provided. To further assess the reliability of the explanations, we compare the results of the proposed framework with three established post-hoc explainability methods (Kernel SHAP, occlusion sensitivity, and Grad-CAM) applied to an independently trained neural network for the function-to-function mapping task. For visualization purposes, the test samples are ordered chronologically in order to preserve the underlying temporal evolution in the presented flow fields and predictions.}

\edit{Table~\ref{tab:accuracy} compares the prediction errors of the proposed XAI model and the FNO baseline across all mapping tasks. Overall, FNO achieves lower mean absolute error (Mean AE) and maximum absolute error (Max AE) values, reflecting its higher expressive capacity as a black-box neural operator. For the blood flow function-to-scalar tasks, the FNO reduces the Mean AE by approximately one order of magnitude compared to the XAI model (e.g., $0.0050 \pm 0.0011$ versus $0.0288 \pm 0.0038$ for Max WSS). A similar trend is observed for the airfoil cases, where the FNO achieves substantially lower errors, particularly at the higher pitching frequency ($f_p = 0.2$), where the more complex oscillatory dynamics pose a greater challenge for the XAI model. In the function-to-function mapping, the FNO achieves a notably lower Mean AE ($0.0087 \pm 0.0017$ versus $0.0733 \pm 0.0018$), although the Max AE values of the two models are closer ($0.5241 \pm 0.1079$ versus $0.5620 \pm 0.0099$), suggesting that both struggle to resolve localized peak values. The overall superiority of FNO should not be a surprise and is well expected (interpretability vs.\ accuracy tradeoff). Importantly, despite its lower numerical accuracy (around one order of magnitude in most cases), the XAI model still captures the dominant qualitative patterns in the data, which is the essential prerequisite for the explainability analysis presented in the following sections. 
}

\begin{table}[h]
    \captionsetup{labelfont={color=black}}
    \caption{\edit{Accuracy comparison between the XAI model (gFLM) and the FNO baseline. Mean Absolute Error (Mean AE) and Maximum Absolute Error (Max AE) are reported on the test dataset across five random seeds.}}
    \label{tab:accuracy}
    \centering
    \small
    \setlength{\tabcolsep}{4pt}
    \renewcommand{\arraystretch}{1.15}
    
    \edit{
    \begin{tabular}{lllcc}
    \toprule
    \textbf{Test case} & \textbf{Task} & \textbf{Model} & \textbf{Mean AE} & \textbf{Max AE} \\
    \midrule
    \multirow{8}{*}{Blood flow} 
        & \multirow{2}{*}{Max WSS}              & gFLM & $0.0288 \pm 0.0038$ & $0.1474 \pm 0.0304$ \\
        &                                       & FNO  & $0.0050 \pm 0.0011$ & $0.0406 \pm 0.0123$ \\
    \cmidrule(lr){2-5}
        & \multirow{2}{*}{Mean WSS}             & gFLM & $0.0239 \pm 0.0025$ & $0.1100 \pm 0.0081$ \\
        &                                       & FNO  & $0.0018 \pm 0.0008$ & $0.0088 \pm 0.0039$ \\
    \cmidrule(lr){2-5}
        & \multirow{2}{*}{Mean DNWSS}           & gFLM & $0.0238 \pm 0.0022$ & $0.1454 \pm 0.0316$ \\
        &                                       & FNO  & $0.0070 \pm 0.0029$ & $0.0876 \pm 0.0497$ \\
    \cmidrule(lr){2-5}
        & \multirow{2}{*}{Function-to-Function} & gFLM & $0.0733 \pm 0.0018$ & $0.5620 \pm 0.0099$ \\
        &                                       & FNO  & $0.0087 \pm 0.0017$ & $0.5241 \pm 0.1079$ \\
    \midrule
    \multirow{8}{*}{Airfoil} 
        & \multirow{2}{*}{$f_p=0.05$, $C_d$}    & gFLM & $0.0032 \pm 0.0003$ & $0.0150 \pm 0.0024$ \\
        &                                       & FNO  & $0.0006 \pm 0.0002$ & $0.0018 \pm 0.0006$ \\
    \cmidrule(lr){2-5}
        & \multirow{2}{*}{$f_p=0.05$, $C_l$}    & gFLM & $0.0047 \pm 0.0003$ & $0.0111 \pm 0.0005$ \\
        &                                       & FNO  & $0.0007 \pm 0.0001$ & $0.0026 \pm 0.0007$ \\
    \cmidrule(lr){2-5}
        & \multirow{2}{*}{$f_p=0.2$, $C_d$}     & gFLM & $0.0712 \pm 0.0065$ & $0.1941 \pm 0.0151$ \\
        &                                       & FNO  & $0.0005 \pm 0.0001$ & $0.0021 \pm 0.0004$ \\
    \cmidrule(lr){2-5}
        & \multirow{2}{*}{$f_p=0.2$, $C_l$}     & gFLM & $0.0858 \pm 0.0074$ & $0.2255 \pm 0.0178$ \\
        &                                       & FNO  & $0.0005 \pm 0.0003$ & $0.0018 \pm 0.0010$ \\
    \bottomrule
    \end{tabular}
    }
    
\end{table}

\subsection{Blood Flow Test Case}
\label{sec:subsection}
\edit{\subsubsection{Function-to-Scalar Mapping}}

Fig.~\ref{fig:ImgToS-Aneurysm} presents the results of the function-to-scalar mapping task for predicting the maximum WSS in the blood flow test case. Fig.~\ref{fig:ImgToS-Aneurysm}a shows representative velocity magnitude fields at selected time steps spanning the pulsatile flow cycle. The input domain is partitioned into four equal subregions to evaluate the contribution of different spatial regions to the model prediction, as illustrated for one representative sample in Fig.~\ref{fig:ImgToS-Aneurysm}b (Sample 1 from Fig.~\ref{fig:ImgToS-Aneurysm}c). \edit{Fig.~\ref{fig:ImgToS-Aneurysm}c compares the XAI model prediction (red dashed line with red square markers) with the ground truth data (black line with black dot markers) across all time steps.} The temporal variation of the maximum WSS generally follows the pulsatile inlet waveform~\cite{csala2022comparing}, and the XAI model reproduces the dominant temporal behavior with good accuracy. \edit{The zonal contributions for the four subregions are shown separately in the four subfigures of Fig.~\ref{fig:ImgToS-Aneurysm}d.} To make the comparison between each zone’s contribution to the XAI model prediction easier, we calculated the mean-centered XAI model prediction and the ground truth output. Pearson correlation coefficients (PC) are computed between each zonal contribution and the overall XAI model prediction, and the values are summarized in Table~\ref{tab:corr}. Among the four regions, Zone~3 exhibits the highest correlation with the XAI model prediction, indicating that this region contributes most strongly to the temporal evolution of the maximum WSS.

\edit{
\begin{figure}[h!]
\centering
\includegraphics[scale=.21]{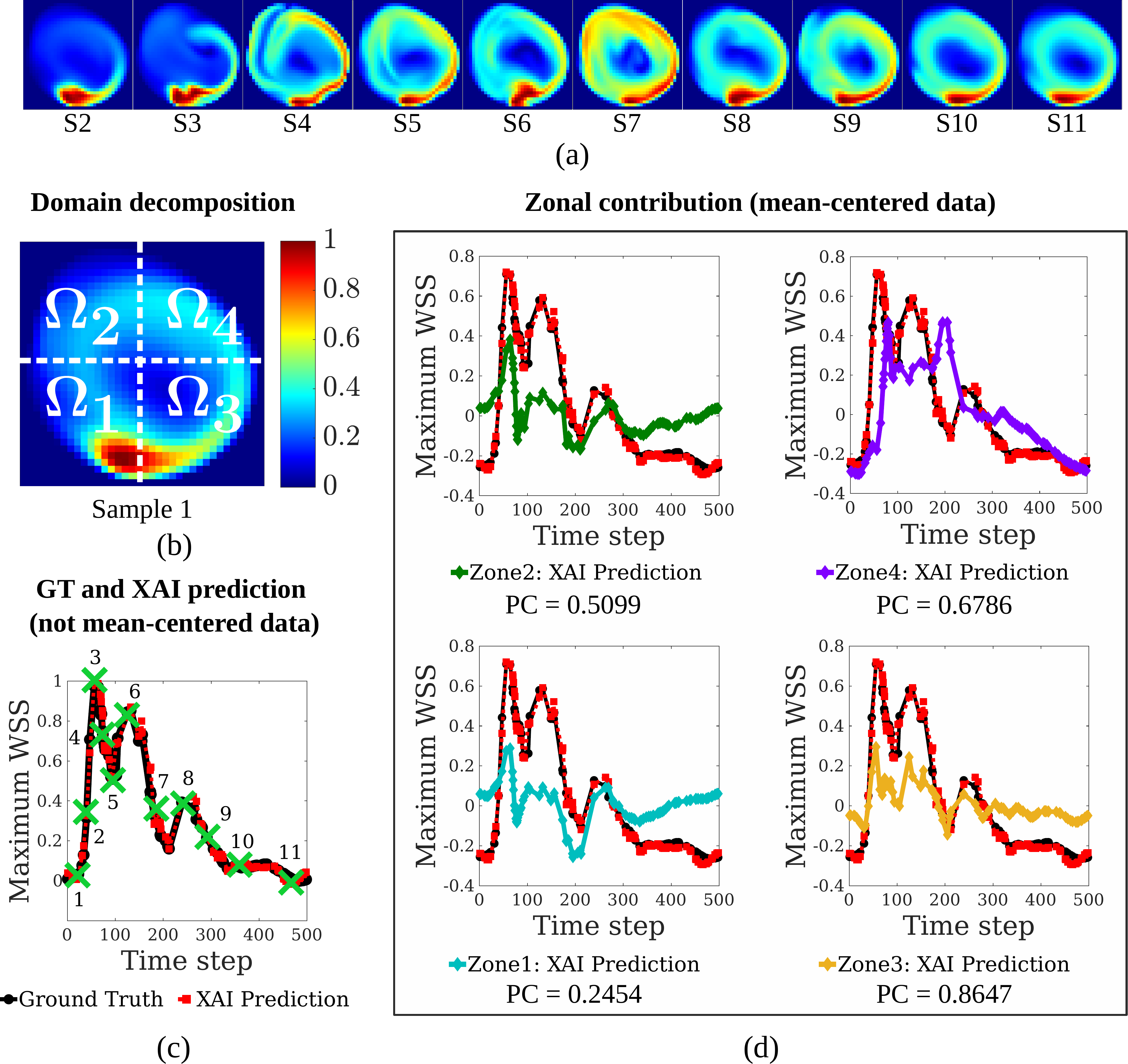}
\caption{\edit{Results of the function-to-scalar mapping task for predicting maximum WSS in the blood flow test case. (a) Velocity magnitude fields at selected test samples (S2--S11) spanning the pulsatile flow cycle. (b) Representative input velocity magnitude field (Sample S1) together with the four-zone domain decomposition used for the zonal contribution analysis.  (c) Ground truth (black) and XAI model prediction (red) of normalized maximum WSS across all test samples, sorted chronologically. Selected samples (S1--S11) are marked on the time-series plot. (d) Mean-centered zonal contributions from each of the four input subregions, with Pearson correlation coefficients (PC) indicated for each zone.}}
\label{fig:ImgToS-Aneurysm}
\end{figure}
}

\begin{table}[h]
    \centering
    \small
    \setlength{\tabcolsep}{1pt}
    \renewcommand{\arraystretch}{1.2}
    \caption{Pearson correlation coefficients between zone-wise contributions and total XAI model prediction for the blood flow and airfoil test cases.}
    \label{tab:corr}
    \begin{tabular}{llcccc}
    \toprule
    \textbf{Test case} & \textbf{Task} & \textbf{Zone 1} & \textbf{Zone 2} & \textbf{Zone 3} & \textbf{Zone 4} \\
    \midrule
    \multirow{4}{*}{Blood Flow} 
        & Max WSS              & \edit{$0.4123 \pm 0.0952$} & \edit{$0.5862 \pm 0.0713$} & \edit{$0.8407 \pm 0.0143$} & \edit{$0.5614 \pm 0.0796$} \\
        & Mean WSS             & \edit{$0.8520 \pm 0.0284$} & \edit{$0.8191 \pm 0.0214$} & \edit{$0.0008 \pm 0.0880$} & \edit{$0.7404 \pm 0.0477$} \\
        & Mean DNWSS           & \edit{$0.3139 \pm 0.1828$} & \edit{$0.9593 \pm 0.0042$} & \edit{$0.7431 \pm 0.0520$} & \edit{$0.4183 \pm 0.0612$} \\
        & Function-to-Function & \edit{$-0.0224 \pm 0.0282$} & \edit{$0.1701 \pm 0.0451$} & \edit{$0.3085 \pm 0.0144$} & \edit{$0.2504 \pm 0.0218$} \\
    \midrule
    \multirow{4}{*}{Airfoil} 
        & $f_p=0.05$, $C_d$    & \edit{$0.8834 \pm 0.0122$} & \edit{$-0.6349 \pm 0.0378$} & \edit{$-0.6385 \pm 0.0240$} & \edit{$0.8790 \pm 0.0108$} \\
        & $f_p=0.05$, $C_l$    & \edit{$-0.9130 \pm 0.0142$} & \edit{$0.9918 \pm 0.0008$} & \edit{$-0.9674 \pm 0.0034$} & \edit{$0.8767 \pm 0.0141$} \\
        & $f_p=0.2$, $C_d$     & \edit{$-0.7956 \pm 0.0193$} & \edit{$0.6772 \pm 0.0233$} & \edit{$-0.2464 \pm 0.1061$} & \edit{$0.8486 \pm 0.0160$} \\
        & $f_p=0.2$, $C_l$     & \edit{$-0.7983 \pm 0.0205$} & \edit{$-0.1200 \pm 0.2149$} & \edit{$0.7890 \pm 0.0209$} & \edit{$0.8560 \pm 0.0175$} \\
    \bottomrule
    \end{tabular}
\end{table}

\begin{figure}[h!]
\centering
\includegraphics[scale=.22]{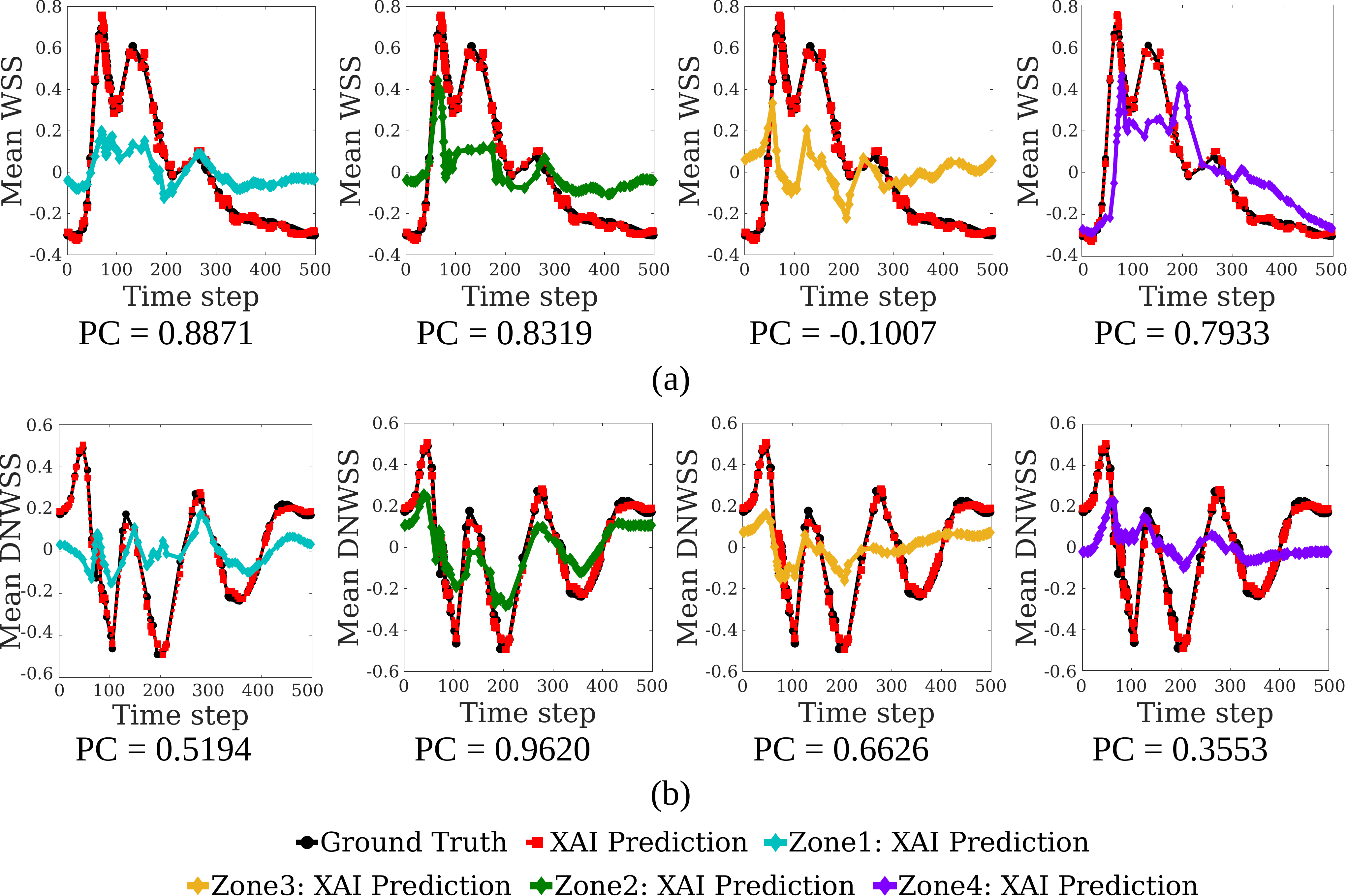}
\caption{\edit{Results of the function-to-scalar mapping task for (a) spatially averaged WSS (Mean WSS) and (b) spatial average of divergence of normalized WSS (Mean DNWSS) in the blood flow test case. Each panel shows the mean-centered zonal contributions from the four input subregions alongside the mean-centered ground truth (black) and XAI model prediction (red). Pearson correlation coefficients (PC) are indicated for each zone.}}
\label{fig:Mean_WSS_DNWSS}
\end{figure}

\edit{Fig.~\ref{fig:Mean_WSS_DNWSS} presents the zonal contribution 
analysis for the spatially averaged WSS (Mean WSS) (Fig.~\ref{fig:Mean_WSS_DNWSS}a) and the spatial average of divergence of normalized WSS (Mean DNWSS) (Fig.~\ref{fig:Mean_WSS_DNWSS}b).} Similar to the maximum WSS case, the XAI model predictions closely follow the ground truth data across the pulsatile cycle. For the spatially averaged WSS, Zone~1 exhibits the highest correlation with the XAI model prediction (\edit{PC $= 0.8871$}), followed closely by Zone~2 (\edit{PC $= 0.8319$}). Zone~3, which dominated the maximum WSS prediction, shows a near-zero correlation (\edit{PC $= -0.1007$}) for this quantity, indicating that different output metrics are driven by different spatial regions of the input domain. \edit{These results suggest that the spatially averaged WSS is governed more strongly by broader flow structures distributed across the lower portion of the aneurysm rather than by the localized high-gradient region associated with the maximum WSS.}

The divergence of the normalized WSS quantifies the spatial variation of the normalized WSS vector field on the projected wall, where positive values correspond to spreading WSS vectors and negative values indicate converging patterns. Fig.~\ref{fig:Mean_WSS_DNWSS}b shows the mean-centered temporal evolution of the spatially averaged absolute DNWSS together with the zonal contributions. \edit{Unlike the pattern observed for maximum WSS, where Zone~3 dominated, here Zone~2 shows the highest correlation (PC $= 0.9620$), aligning closely with the ground truth and XAI model prediction across most time steps. This shift in the dominant zone across different output quantities suggests that the integral operator adapts to the specific spatial features relevant to each physical metric, rather than relying on a single input region for all predictions.}

The XAI model associates Zone~3 with the maximum WSS and Zone~1 with the spatially averaged WSS as the dominant contributing regions. These zones exhibit large variations in velocity magnitude across their spatial extent, giving rise to strong velocity gradients. This observation is consistent with the known physical relationship between velocity gradients and wall shear stress. \edit{For the spatial average of the divergence of normalized WSS, Zone~2 emerges as the most influential region, suggesting that flow structures driving WSS topology differ from those governing WSS magnitude itself.} Overall, the results indicate that regions with higher variability in the velocity field, particularly within a plane away from the wall, play a significant role in shaping different aspects of the WSS response. Importantly, the XAI model identifies these physically meaningful regions without any prior physical constraints, providing an interpretable view of how different parts of the input contribute to different characteristics of the output. \edit{An analysis of how combinations of zonal contributions approximate the overall XAI model prediction is presented in Appendix (Sec.~\ref{sec:sum_contributions}).}

\edit{\subsubsection{Function-to-Function Mapping}}

\begin{figure}[h!]
\centering
\includegraphics[scale=0.32]{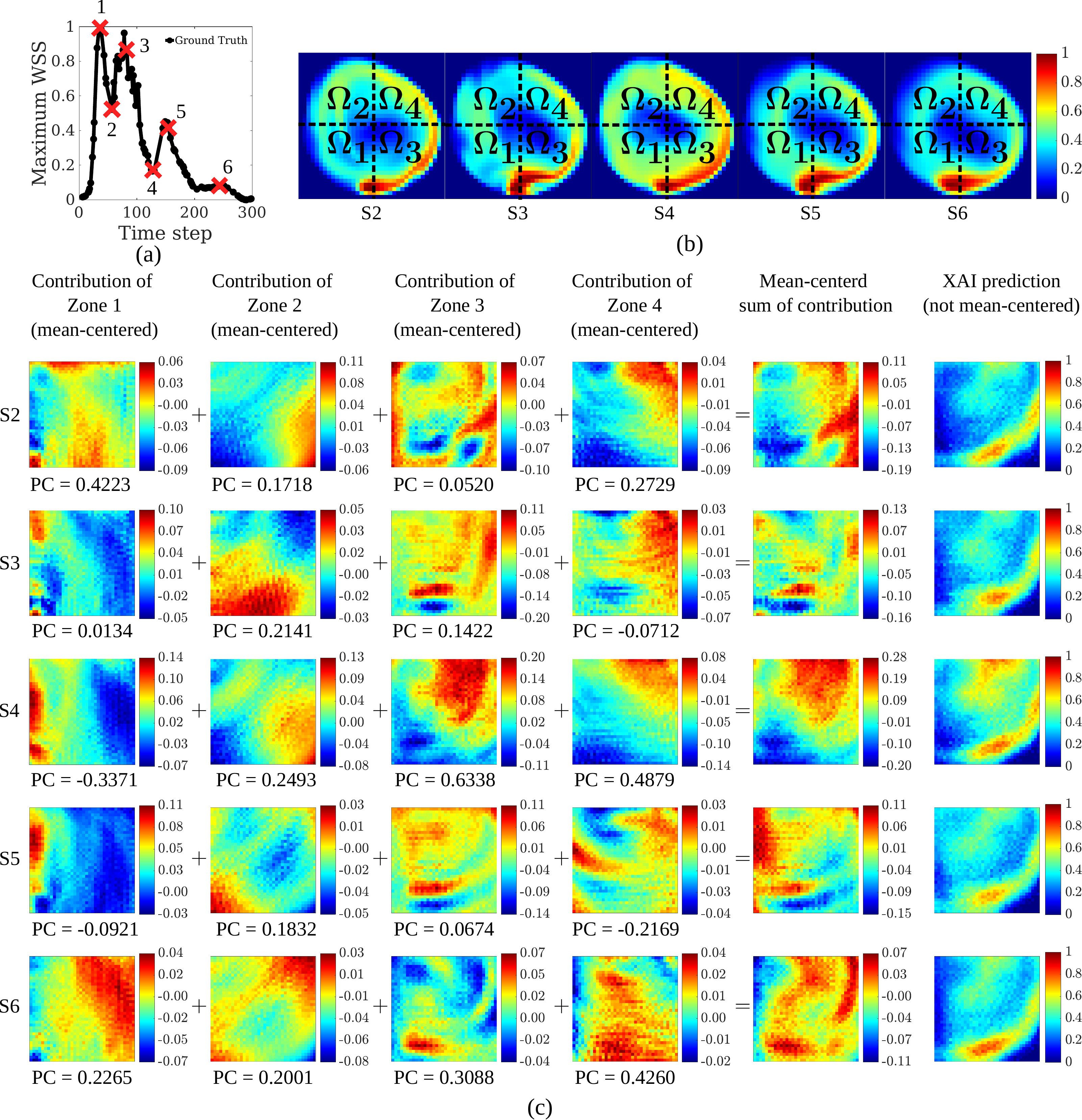}
\caption{\edit{Results of the function-to-function mapping task for the blood flow test case. (a) Maximum WSS versus time step for the test dataset, with six representative samples (S2--S6) marked. (b) Velocity magnitude fields at the corresponding samples, with the four-zone input domain decomposition indicated. (c) Mean-centered zonal contributions from each input subregion at the selected samples, alongside the mean-centered sum of contributions and the full XAI model prediction (not mean-centered). Pearson correlation coefficients (PC) between each zone's contribution and the total XAI prediction are shown for each sample.}}
\label{fig:F2F_ZonesCont}
\end{figure}

In the function-to-function mapping task, the XAI model achieved higher absolute errors (Table~\ref{tab:accuracy}); however, the model still provides predictions with good qualitative accuracy, effectively capturing major spatial patterns present in the data. Considering the plot of the maximum WSS across time steps (Fig.~\ref{fig:F2F_ZonesCont}a), five representative time steps are selected for a detailed analysis. Fig.~\ref{fig:F2F_ZonesCont}b displays the velocity magnitude fields at these selected time steps, showing that high velocities mainly occur within Zone 3 of the input domain. These input functions are used to predict the WSS function in the region with localized high WSS (Fig.~\ref{fig:Schematic}). Fig.~\ref{fig:F2F_ZonesCont}c presents the individual contributions of each of the four zones alongside the total XAI model prediction at corresponding time steps. From Fig.~\ref{fig:F2F_ZonesCont}c, by qualitatively comparing the zonal contributions with total XAI model prediction, it can be concluded that the output of the XAI model from Zone 3 (mean-subtracted) consistently dominates the main patterns observed in mean-subtracted XAI model prediction (sum of contributions), by showing distinct positive and negative contributions that approximately highlight patterns corresponding to high and low WSS values, respectively. The results of Pearson correlation calculations, represented in Table~\ref{tab:corr}, also indicate that the average correlation coefficient between Zone 3's contribution and XAI model prediction has the highest value. \edit{It is worth noting that at individual time steps, Zone~3 does not always exhibit the highest correlation among the four zones (e.g., PC $= 0.0520$ at S2); however, its average correlation across the entire test dataset is consistently the highest, indicating that the dominance of this region is a robust trend rather than a snapshot-specific observation.} The consistency of this observation across time steps confirms that for the task of mapping the velocity magnitude field to the localized high WSS region, the velocity features in Zone 3 primarily govern the overall output pattern. Conversely, no consistent pattern is observed for the other three zones' outputs across the selected time steps.

\edit{\paragraph{Combination of Zonal Contributions}}

\begin{figure}[h!]
\centering
\includegraphics[scale=.21]{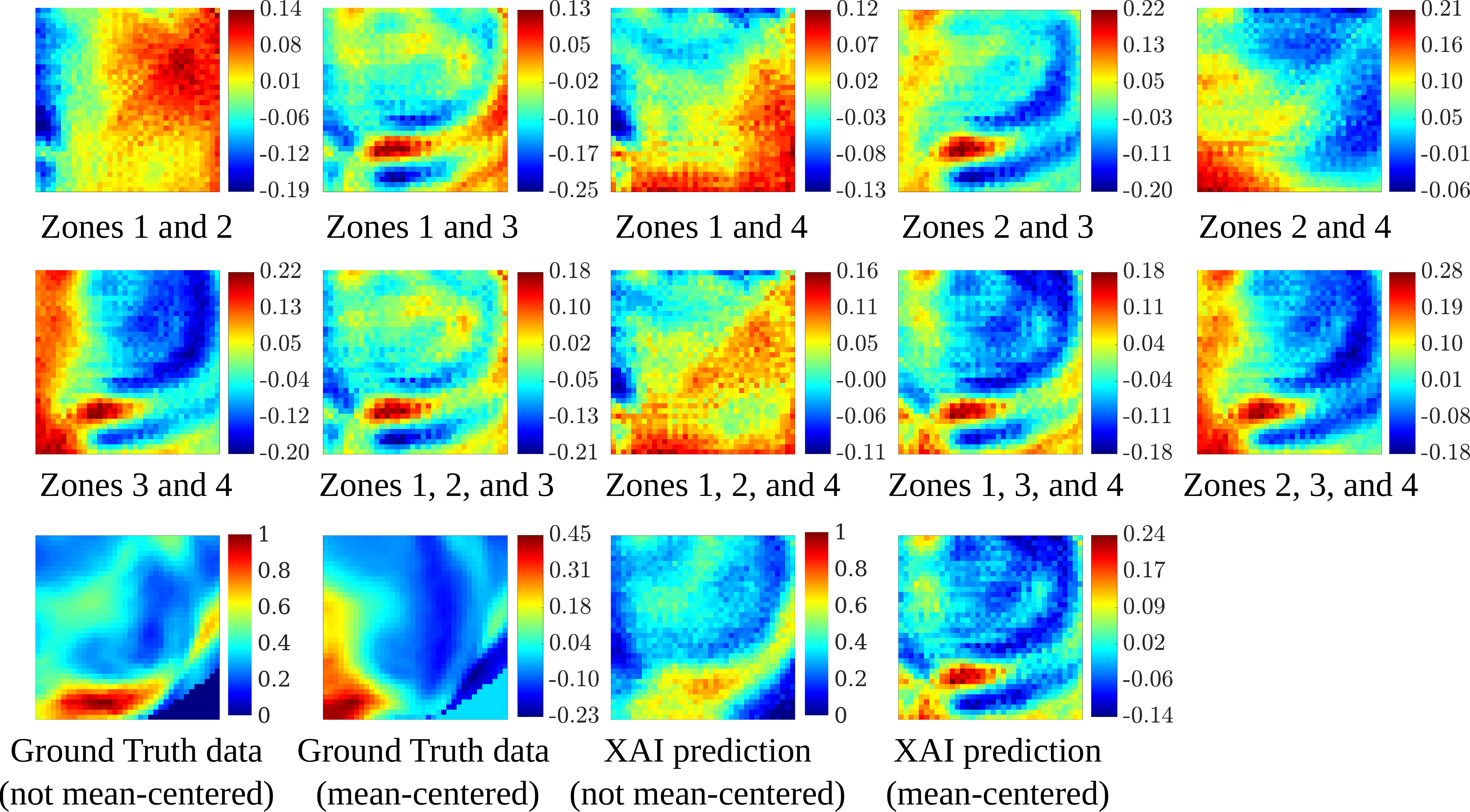}
\caption{\edit{Combinations of zonal contributions for the function-to-function mapping task at Sample 1 (marked in Fig.~\ref{fig:F2F_ZonesCont}a). The first two rows show all possible two-zone and three-zone combinations of mean-centered zonal contributions. The bottom row shows the ground truth (not mean-centered and mean-centered) and the XAI model prediction (not mean-centered and mean-centered) for reference.}}
\label{fig:Sum_Zones_38}
\end{figure}

Fig.~\ref{fig:Sum_Zones_38} demonstrates the impact of different zonal contribution combinations on the XAI model predictions at a representative time step. As shown in Fig.~\ref{fig:Sum_Zones_38}, combining the contribution from Zone 3 with other zones yields results most closely resembling the overall XAI model prediction (mean-centered). Specifically, adding the contribution of Zone 2 to that of Zone 3 (Zone 2 + Zone 3 contributions) enhances the alignment of the outcome with the target prediction pattern both quantitatively and qualitatively. In contrast, combinations excluding Zone 3's contribution fail to capture essential characteristics of the overall prediction. This reinforces our previous function-to-scalar conclusion that Zone 3 is the most influential region in producing maximum WSS as the single scalar output, capturing significant spatial patterns in the input function domain and dominating the predictive output function.

\edit{\paragraph{Role of Non-linearity}}

\edit{An additional analysis examining the effect of introducing nonlinear integral terms into the candidate library is presented in Appendix (Sec.~\ref{sec:nonlinear}). The results show that, although nonlinear terms slightly improved the Mean AE, they also increased the numerical ill-conditioning of the regression problem and did not lead to a substantial improvement in the overall predictive performance.}

\edit{\paragraph{Time-Varying Analysis}}

\edit{A time-varying analysis converting the function-to-function outputs into spatially averaged scalar values is presented in Appendix (Sec.~\ref{sec:f2f_timevarying}), confirming that Zone~3 remains the dominant contributor across the full test dataset.}

\edit{\paragraph{Comparison with Post-hoc Explainability methods}}

\begin{figure}[h!]
\centering
\includegraphics[scale=.23]{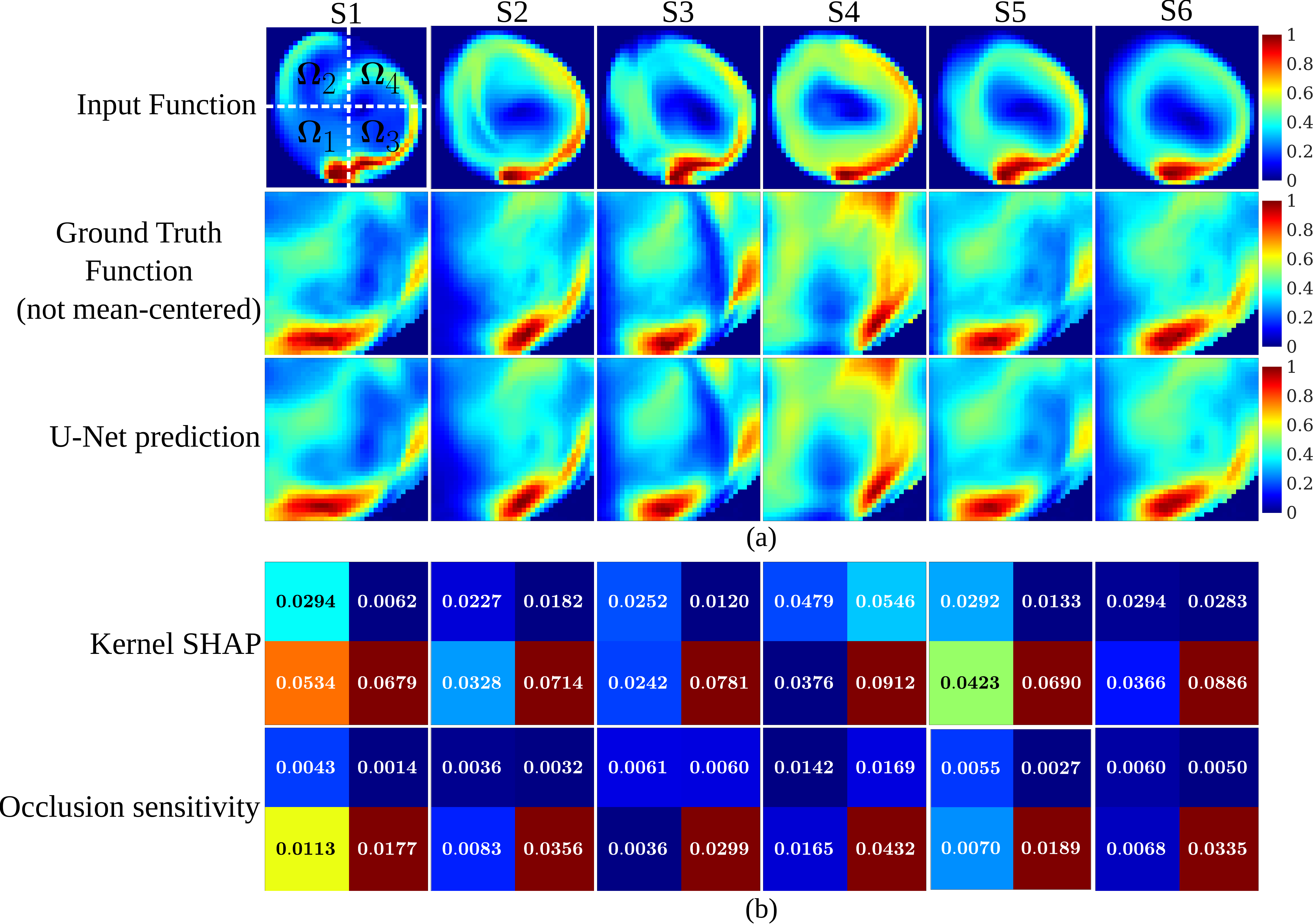}
\caption{\edit{Comparison of post-hoc explainability methods for the function-to-function mapping task in the blood flow test case. (a) Input velocity fields, ground truth WSS outputs, and U-Net predictions for six representative test samples (S1--S6). The four-zone input domain decomposition is indicated on sample S1. (b) Zonal importance scores from Kernel SHAP (top row) and occlusion sensitivity (bottom row). Each $2 \times 2$ block corresponds to the four input zones, with the importance score displayed numerically within each cell. Higher values (dark red zones) indicate greater influence on the prediction.}}
\label{fig:shap}
\end{figure}

\edit{To validate the explanations produced by the proposed framework, we compared the zonal importance rankings with those obtained from two independent post-hoc explainability methods (Kernel SHAP and occlusion sensitivity) applied to a separately trained U-Net model on the same function-to-function mapping task.} It should be noted that the XAI model (integral operator) was trained directly on the simulation data without any neural network; the U-Net serves solely as a black-box reference model for the post-hoc comparison. Fig.~\ref{fig:shap}a shows input velocity magnitude fields together with the corresponding ground truth and U-Net-predicted output functions at six representative test samples (S1--S6). The U-Net predictions closely reproduce the dominant spatial patterns observed in the ground truth data.

\edit{Fig.~\ref{fig:shap}b presents the zonal importance scores computed by Kernel SHAP (top) and occlusion sensitivity (bottom) for the same samples. The numerical importance values are reported within each cell.} In Kernel SHAP, each zone's importance is quantified by its absolute SHAP value, where larger values indicate a stronger influence on the prediction. \edit{In the occlusion sensitivity analysis, each zone is individually masked with zeros, and the resulting change in prediction error is recorded; a larger change indicates greater dependence on that zone.}

\edit{For both SHAP and occlusion sensitivity, Zone~3 consistently exhibits the largest importance values across test samples, indicating that the lower-right region of the input domain contributes most strongly to the prediction.} This observation is consistent with the zonal contribution analysis of the proposed XAI framework, where Zone~3 also showed the highest Pearson correlation with the overall model prediction (Table~\ref{tab:corr}). \edit{When the importance scores are averaged across all test samples, both Kernel SHAP and occlusion sensitivity rank Zone~1 as the second most important zone and Zone~4 as the least influential.} The XAI model's Pearson correlation analysis, by contrast, places Zone~4 second and Zone~1 lowest. \edit{While the three methods agree on the dominant region, these differences in the ordering of lower-ranked zones reflect the fundamentally different mechanisms through which each method quantifies importance: perturbation-based approaches (Kernel SHAP and occlusion sensitivity) measure how much the prediction changes when a zone is removed, whereas the XAI model's correlation metric captures how closely a zone's contribution pattern tracks the overall output over time. Despite these differences, the consistent identification of Zone~3 as the primary driver of the WSS prediction across all three independent approaches provides cross-validation of the proposed framework's explanations. A further comparison using Grad-CAM is presented in Appendix (Sec.~\ref{sec:gradcam}).}

\subsection{Airfoil Test Case}
\label{sec:subsection}

\edit{\subsubsection{Low-frequency case (\boldmath{$f_p = 0.05$})}}

\begin{figure}[h!]
\centering
\includegraphics[scale=.25]{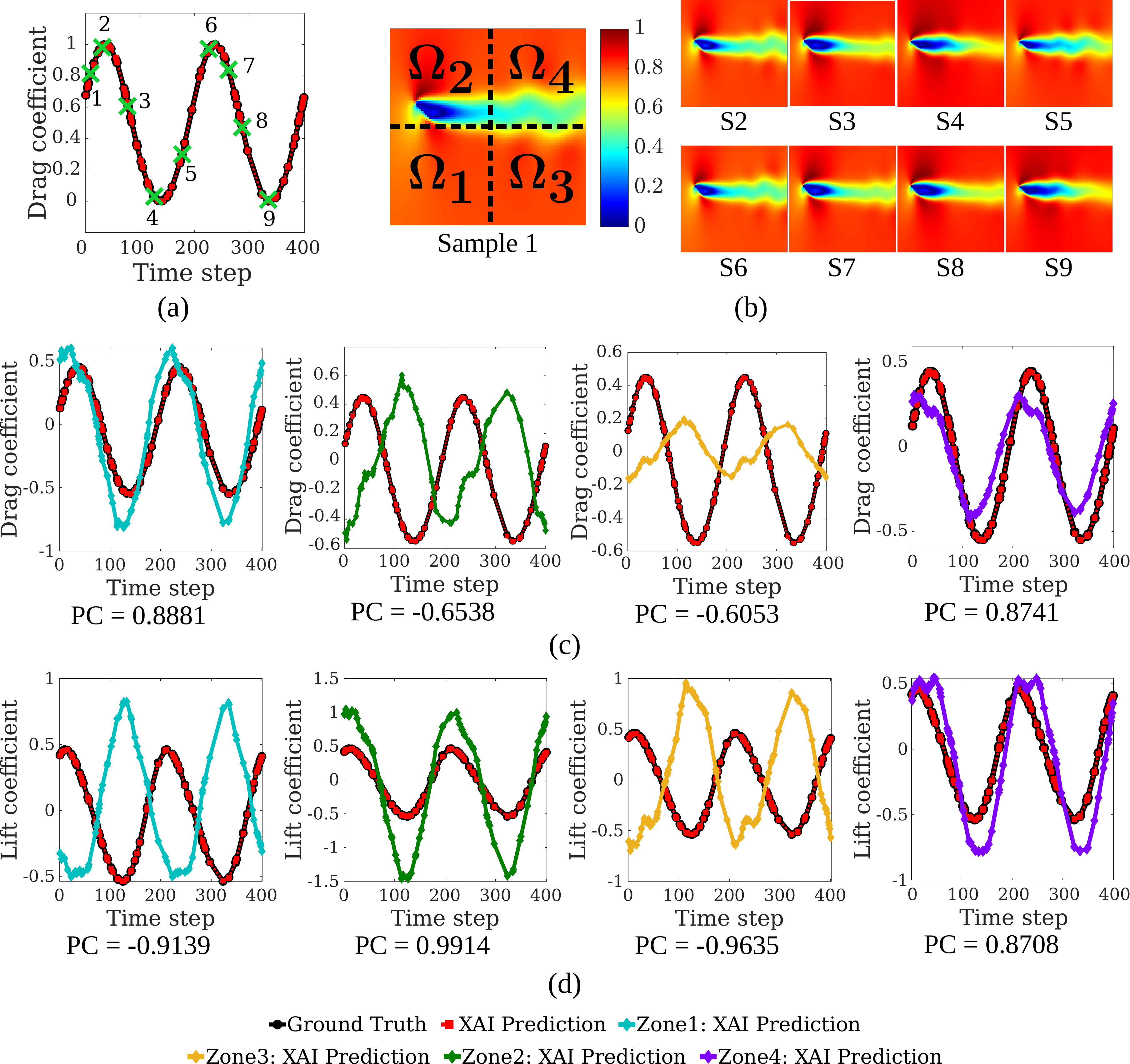}
\caption{\edit{Results of the function-to-scalar mapping task for the airfoil test case at nondimensional pitching frequency $f_p = 0.05$. (a) Time variation of the drag coefficient with selected representative samples highlighted along the cycle. (b) Velocity magnitude fields corresponding to the selected time steps, together with the four-zone decomposition of the input domain. (c) Mean-centered zonal contributions for the drag coefficient prediction compared with the mean-centered ground truth data and the mean-centered XAI model prediction. (d) Mean-centered zonal contributions for the lift coefficient prediction. The Pearson correlation coefficients (PC) between each zonal contribution and the overall XAI model prediction are reported within the plots.}}
\label{fig:Airfoil05}
\end{figure}

Fig.~\ref{fig:Airfoil05} illustrates the results obtained from mapping the velocity magnitude field over an airfoil to scalar outputs, namely the drag and lift coefficients, at a nondimensional pitching frequency of $f_p = 0.05$. The velocity magnitude fields at selected time steps (Figs.~\ref{fig:Airfoil05}a and ~\ref{fig:Airfoil05}b) show that the flow remains relatively steady without strong oscillations at this frequency, with the minimum velocities occurring predominantly within the separation region behind the airfoil.

Figs.~\ref{fig:Airfoil05}c and~\ref{fig:Airfoil05}d present the time-series results for the drag and lift coefficients, respectively. For the drag coefficient prediction (Fig.~\ref{fig:Airfoil05}c), the XAI model predictions (red square markers) closely follow the ground truth data (black markers), demonstrating good agreement throughout the cycle. \edit{The corresponding zonal contributions are also shown after mean-centering in order to better highlight the temporal relationship between each zone and the overall prediction. The Pearson correlation coefficients (PC) between each zonal contribution and the overall XAI model prediction are reported directly in the figure and summarized quantitatively in Table~\ref{tab:corr}.} According to these results, Zone~1 exhibits the strongest positive correlation with the drag prediction, closely followed by Zone~4, whereas Zones~2 and~3 show strong negative correlations. \edit{These results suggest that the drag response is primarily influenced by flow structures in the lower-left region of the domain, as well as by the wake region behind the airfoil associated with Zone~4.}

A similar analysis was performed for the lift coefficient (Fig.~\ref{fig:Airfoil05}d). In this case, however, the dominant contribution shifts to Zone~2, which exhibits an exceptionally strong positive correlation with the lift prediction (\edit{PC = $0.9914$}). This suggests that the lift coefficient is governed more strongly by the upper flow structures surrounding the airfoil, particularly those associated with near-wall flow variations captured within Zone~2. \edit{The strong negative correlations observed for Zones~1 and~3 indicate that these regions contribute to the lift dynamics in a manner that systematically opposes the dominant oscillatory trend of the prediction. In other words, when the contribution from Zone~2 acts to increase the lift coefficient, the contributions from Zones~1 and~3 tend to vary in the opposite direction. This behavior suggests the presence of competing flow mechanisms within different regions of the domain, reflecting the asymmetric interaction between the upper and lower flow structures surrounding the airfoil during the oscillation cycle.}

\edit{\subsubsection{High-frequency case (\boldmath{$f_p = 0.2$})}}

\begin{figure}[h!]
\centering
\includegraphics[scale=.25]{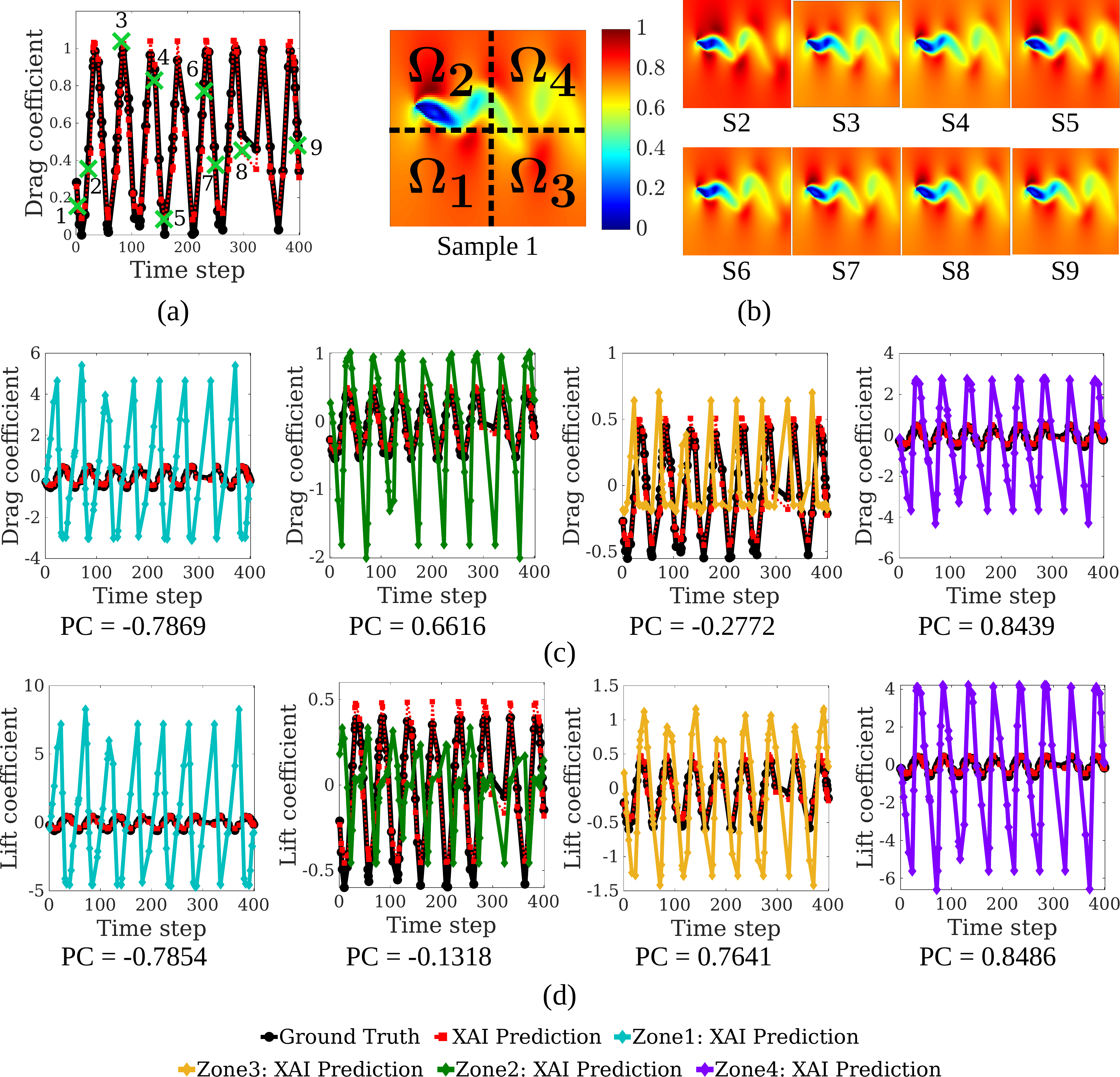}
\caption{\edit{Results of the function-to-scalar mapping task for the airfoil test case at nondimensional pitching frequency $f_p = 0.2$. (a) Time variation of the drag coefficient with selected representative samples highlighted along the cycle. (b) Velocity magnitude fields corresponding to the selected time steps, together with the four-zone decomposition of the input domain. (c) Mean-centered zonal contributions for the drag coefficient prediction compared with the mean-centered ground truth data and the mean-centered XAI model prediction. (d) Mean-centered zonal contributions for the lift coefficient prediction. The Pearson correlation coefficients (PC) between each zonal contribution and the overall XAI model prediction are reported within the plots.}}
\label{fig:Airfoil2}
\end{figure}

We next reevaluate the function-to-scalar mapping task for the airfoil case at a higher nondimensional pitching frequency of $f_p = 0.2$, where the flow over the airfoil exhibits substantially stronger oscillatory behavior. Figs.~\ref{fig:Airfoil2}a and~\ref{fig:Airfoil2}b present representative velocity magnitude fields illustrating the evolving wake structures and increased flow unsteadiness at this frequency. Figs.~\ref{fig:Airfoil2}c and~\ref{fig:Airfoil2}d show the corresponding time-series results for the drag and lift coefficients, respectively. Compared with the lower-frequency case, both coefficients now exhibit more rapid oscillations and larger temporal variations. For the drag coefficient prediction (Fig.~\ref{fig:Airfoil2}c), Zone~4 exhibits the strongest positive correlation with the XAI model prediction, while Zone~1 shows the strongest negative correlation. Zones~2 and~3 contribute less strongly, although Zone~2 remains positively correlated with the prediction. \edit{A similar trend is observed for the lift coefficient (Fig.~\ref{fig:Airfoil2}d), where Zone~4 again shows the strongest positive correlation, followed by Zone~3, whereas Zones~1 and~2 exhibit negative correlations.} 

\edit{Compared with the $f_p = 0.05$ case, the dominant regions shift from near-wall and localized flow structures toward larger-scale wake dynamics, reflecting that the aerodynamic force predictions become increasingly dominated by flow structures associated with the wake and vortex shedding region located in the upper-right portion of the domain (Zone~4). Furthermore, the strong negative correlations observed for Zones~1 indicate that this region contributes to the aerodynamic response in opposition to the dominant wake-driven dynamics associated with Zones~4. Physically, this suggests that different regions of the flow field act to either reinforce or counterbalance the oscillatory aerodynamic forces throughout the pitching cycle. Similar to the blood flow test case, an analysis of how combinations of zonal contributions approximate the overall XAI model prediction for the airfoil cases is provided in Appendix (Sec.~\ref{sec:sum_contributions}).}

\section{Discussion} \label{sec:disc}

Our proposed XAI framework leverages kernel-based integral equations to achieve interpretable operator learning, which enables the modeling of relationships between functional data through functional linear models. Kernels encode nonlocal interactions and capture the spatial dependencies critical in operator learning tasks~\cite{liuschiaffini2024neural}. Drawing inspiration from sparse identification of nonlinear dynamics (SINDy)~\cite{brunton2016discovering}, our approach constructs a library of candidate integral equations equipped with diverse kernels. However, unlike SINDy, which prioritizes sparsity to discover governing equations, explaining the patterns in data within our framework does not require a sparse regression model. Additionally, by representing the operator as a linear combination of integral equations, our model mirrors the ensemble strategy of neural additive models, where each integral equation approximates part of the solution, and the final prediction is a linear superposition of these terms~\cite{agarwal2021neural}. This parallel addition of kernels provides analytical simplicity and design transparency, in contrast to neural operators like Fourier Neural Operators (FNO)~\cite{li2020fourier}, where kernels are applied sequentially across hidden layers, reducing interpretability.

\edit{The accuracy comparison in Table~\ref{tab:accuracy} provides quantitative evidence for the accuracy--interpretability trade-off inherent in our framework. Across all tasks, the FNO baseline achieves lower prediction errors, which is expected given its substantially higher number of trainable parameters and nonlinear expressive capacity. However, this gain in accuracy comes at the expense of interpretability, as the FNO operates as an opaque model whose predictions cannot be directly decomposed into physically meaningful contributions without additional post-hoc analysis. In contrast, the XAI model sacrifices some predictive precision in exchange for built-in explainability rooted in its integral equation structure, which directly enables the zonal decomposition analysis reported here. The airfoil cases at $f_p=0.2$ illustrate this trade-off clearly, where the complex oscillatory dynamics increase the XAI model's prediction errors relative to FNO, yet the zonal contribution analysis still pinpoints the wake and vortex shedding region as the dominant driver of its aerodynamic force predictions. We emphasize that the goal of the proposed framework is not to compete with neural operators in predictive accuracy, but rather to provide a self-explainable model that reveals the underlying spatial relationships between input and output patterns.}

While existing operator learning techniques, such as Graph Neural Operators (GNO)~\cite{li2020neural} and kernel/Gaussian-process-based methods~\cite{williams2006gaussian, garrigaalonso2019deep} excel at approximating mappings between infinite-dimensional function spaces, their complex neural network architectures limit transparency and the ability to explicitly encode domain-specific structure. In contrast, the parallel, linear structure of our proposed model enables the zonal decomposition: because the total output is a simple sum of contributions from different input regions, one can quantify each region's role without approximation or external tools. Compared with post-hoc XAI methods such as SHAP~\cite{lundberg2017unified} and LIME~\cite{ribeiro2016should}, which treat the model as a black box and attribute importance only at the input level, our framework embeds explainability into the model itself and extends it to the output level by revealing the spatial pattern each input zone produces in the predicted field. \edit{Post-hoc methods also require recomputation for every new input and can yield inconsistent explanations across different explainers~\cite{turbe2023evaluation, slack2020fooling}, whereas the proposed XAI model's explanations follow directly from its mathematical structure and remain stable by construction. The reduced predictive accuracy relative to neural operators, as quantified in Table~\ref{tab:accuracy}, is the expected cost of this transparency, but for applications where understanding the input--output relationship matters as much as prediction quality, the trade-off is justified. Taken together, these differences highlight that the proposed framework is not simply an extension of existing operator-learning or functional regression approaches, but a reformulation that prioritizes analytical transparency and built-in explainability at the operator level.}



By quantifying correlations between individual zonal contributions and the overall prediction, we can identify which input features are most influential during the probing phase of the integral operator. Specifically, in most test cases, the operator adapts to the existing relationships between inputs and outputs by prioritizing regions with the highest feature gradients. Therefore, our integral operator inherently explains its decision-making process through a mathematically interpretable framework. The self-explanatory nature of the integral operator distinguishes our approach from other post-hoc explainable methods, which often rely on algorithms that lack inherent interpretability and internal transparency~\cite{rudin2019stop, bhalla2023discriminative, lundberg2017unified}. That is, one could argue that many post-hoc XAI models themselves are not very transparent. Furthermore, unlike these conventional methods, which typically require a pretrained neural network, our model is inherently data-driven, and in this work, was directly applied to the training data without any neural network. However, as demonstrated in our prior work~\cite{arzani2024interpreting}, our approach can also be applied to interpret pretrained neural networks, offering versatility across both data-driven and neural network-driven contexts.

Analysis of the combination plots reveals that information from at least two zones must be summed up to generate a pattern with strong correlation to the overall XAI prediction. In some test cases, these combinations exclude zones that individually exhibit high correlation with the XAI model predictions (e.g., maximum WSS and spatially averaged WSS). This suggests that the specific features of these zones have a distinct influence on the predictions, and excluding them forces the model to integrate information from additional zones to achieve the same predictions. Conversely, in other function-to-scalar mapping tasks in this study, combining the zone with high individual correlation with other zones leads to a more accurate match with the overall XAI prediction. 

Our framework provides the opportunity to examine which features of the input are most relevant to the underlying physical behavior of the system, leveraging the operator's built-in explainability. The ability of the XAI model to decompose the overall output into constituent contributions provides insights into the spatial features of the input function and how they collectively shape the output. For instance, in the aneurysm case, the strong correlation between Zone 3's contribution and the model predictions reflects the known link between steep velocity gradients and elevated WSS values. \edit{In the airfoil test case, the dominant regions vary depending on the flow regime. At the lower pitching frequency ($f_p = 0.05$), Zone 2, characterized by strong gradients in the velocity field, shows the highest correlation with the lift coefficient, while both Zone 1 and Zone 4 contribute substantially to the drag prediction. At the higher frequency ($f_p = 0.2$), Zone 4 consistently emerges as the most influential region for both drag and lift coefficient predictions. This region corresponds to the wake, where vortex shedding effects are more prominent.} There is no requirement for these predictions to align with known physical principles, and the integral operator might focus on non-trivial features in the data to make its predictions. \edit{Rather than claiming that the model discovers new physics, we interpret the decomposition as a structured diagnostic tool: it can identify spatial relationships that are consistent with established theory, flag unexpected dependencies worth further investigation, or identify which input regions carry the most predictive information.} From a practical perspective, the proposed framework may also provide useful guidance for experimental design. For example, if the decomposition consistently identifies a limited subset of spatial regions as dominant contributors to the prediction, these regions could potentially guide future sensor placement strategies or reduced-order measurement approaches. \edit{However, such applications were not investigated directly in the present study and remain an important direction for future work. We emphasize that the present test cases were selected because the underlying physics is well understood, allowing us to verify that the model's explanations remain physically plausible. Extending the framework to systems where the governing physics is less well characterized is a natural direction for future work.}

\edit{To compare the explanations produced by our framework with established post-hoc XAI approaches, we applied Kernel SHAP, occlusion sensitivity, and Grad-CAM to a neural network (U-Net model) trained on the function-to-function mapping test case. In the present study, the Grad-CAM activation maps from the shallow encoder stages consistently highlighted regions with strong local velocity gradients, particularly within the lower-right portion of the domain, which aligns qualitatively with the dominant regions identified by the proposed XAI framework, Kernel SHAP, and occlusion sensitivity analyses. Despite relying on fundamentally different mechanisms, all three post-hoc methods identified a dominant spatial region that was consistent with the influential zone detected by our integral operator framework.} The novelty of our proposed XAI model lies in its ability not only to detect the most important zone but also to elucidate the role this zone plays in shaping the output spatial pattern. This ability is inherent and can be leveraged through the integral operator established in this work, which captures the input-output spatial dependencies. To the best of our knowledge, this approach is the first to provide explainability at the output level by systematically sorting the decomposed outputs. The linear decomposition could be compared with PCA and proper orthogonal decomposition (POD), commonly used in unsteady fluid mechanics~\cite{csala2022comparing}, where the output patterns are decomposed into linearly additive modes.

\edit{
Although the post-hoc explainability methods investigated in this study (Kernel SHAP, occlusion sensitivity, and Grad-CAM) identified spatially consistent influential regions, it is important to recognize that these methods characterize different aspects of model behavior and model reliance rather than directly measuring the intrinsic information content of the input regions themselves. Grad-CAM is fundamentally gradient-based and highlights regions that most strongly influence the model output through the internal feature representations learned by the network~\cite{selvaraju2020grad}. Consequently, the resulting saliency maps are highly model-dependent and primarily reflect how the trained network processes the input. As a result, informative regions may receive weak attribution if the network has not learned to utilize them effectively, while regions associated with strong gradients or activations may appear highly influential because of architectural or optimization-related biases~\cite{adebayo2018sanity, kindermans2019unreliability}.

Similarly, Kernel SHAP and occlusion sensitivity quantify how the model output changes when portions of the input are perturbed or removed~\cite{petsiuk2018rise}. While this provides a more causal interpretation than purely gradient-based methods, the resulting importance scores still measure the sensitivity of the trained model to perturbations rather than the intrinsic information contained within the data itself. In particular, occlusion-based methods may underestimate the importance of informative regions when redundant information exists elsewhere in the domain, since the model may compensate using alternative spatial features. Conversely, masking operations can introduce artificial distribution shifts that may exaggerate the apparent importance of certain regions~\cite{fong2017interpretable}. These considerations highlight an important distinction between information content and model reliance. Information content is fundamentally a property of the underlying data distribution and reflects what could, in principle, be learned from a region. In contrast, explainability methods quantify what the trained model actually uses during prediction. As a result, different explainability approaches may produce partially inconsistent attribution maps even for the same input sample, since they probe different aspects of model behavior~\cite{samek2021explaining}.

In contrast, the proposed framework generates predictions and explanations simultaneously through the structure of the integral operator itself, without relying on an additional external explainability tool. Because the model is formulated directly in terms of the input data and analytical integral terms, the resulting explanations are less dependent on hidden latent representations or internal feature hierarchies that may emerge during neural network training. This distinction is particularly important in scientific machine learning, where neural networks may preferentially emphasize certain spatial features because of optimization dynamics, spectral bias, or architectural characteristics rather than their true physical importance~\cite{duraisamy2026predictivity}. Nevertheless, the qualitative agreement observed between the proposed explainability framework, Kernel SHAP, Grad-CAM, and occlusion sensitivity provides increased confidence that the identified high-gradient transition regions correspond to physically meaningful structures that are consistently utilized during prediction rather than artifacts specific to a single explainability method or model architecture.

}


\edit{To further examine the applicability of the proposed framework, we also conducted additional experiments on a synthetic dataset generated from a semi-linear elliptic PDE problem~\cite{hasani2025generating}. Further details and representative results are provided in Appendix (Sec.~\ref{sec:newdataset}). In this case, the input functions exhibited highly irregular and spatially chaotic patterns without clearly identifiable coherent structures, while the corresponding output functions remained comparatively smooth. Under these conditions, the resulting zonal contributions became substantially more difficult to interpret physically, with individual regions rarely exhibiting clear or consistent dominant behavior across samples. These observations suggest that the reliability of the decomposition-based explanation is closely tied to the presence of spatially organized or coherent patterns within the input functions.}

\edit{As discussed previously, the proposed framework deliberately adopts a linear decomposition structure to preserve exact additive explainability of the input--output mapping. Additional experiments with nonlinear integral terms, discussed in Appendix (Sec.~\ref{sec:nonlinear}), showed that augmenting the candidate library with nonlinear activations produced only marginal changes in predictive accuracy while substantially increasing the numerical ill-conditioning of the regression problem.} Additionally, the predictive performance of the model depends strongly on the bandwidth range defined in the exponential kernels. Fine-tuning these hyperparameters can become increasingly challenging for complex systems with highly nonlinear or irregular input patterns. Future work will utilize novel global optimization strategies to overcome this challenge. \edit{Finally, although the present study focused exclusively on fluid flow problems, the mathematical structure of the proposed framework is adaptable to other operator-learning settings, such as solid mechanics and heat transfer. Investigating the robustness and effectiveness of the framework in such domains remains an important direction for future work.}

\section{Conclusion} \label{sec:disc}

\edit{In this study, we introduced a self-explainable operator-learning framework based on generalized functional linear models and kernel-based integral equations. By representing the operator as a linear superposition of integral terms, the proposed framework enables exact additive decomposition of predictions into zonal contributions, providing built-in explainability at both the input and output levels. Unlike conventional neural operators that rely on post-hoc methods, the proposed approach embeds explainability directly into the operator structure through a mathematically transparent formulation. The additive decomposition links localized input features to the formation of patterns in the output, enabling the model’s internal reasoning process to be interpreted directly from the learned operator structure.  The framework was evaluated on blood flow and unsteady aerodynamic problems. The resulting zonal decompositions consistently identified physically meaningful structures. Comparisons with Kernel SHAP, occlusion sensitivity, and Grad-CAM showed qualitative agreement in the dominant contributing regions, supporting the consistency of the proposed explainability framework. Although the framework provides exact additive explainability, its predictive accuracy remains lower than that of the Fourier Neural Operator, and the interpretability of the zonal contributions becomes more challenging for highly irregular input patterns. Overall, the proposed framework demonstrates that operator learning can be reformulated to prioritize analytical transparency and interpretable input--output relationships while still capturing meaningful physical behavior. Future work will focus on developing more expressive yet interpretable nonlinear formulations and extending the framework to broader operator-learning settings.
}

\section*{Acknowledgements}
This work was supported by NSF Award No.~2247173.

\section*{Competing Interests}
The authors have no conflicts of interest.

\section{Appendix} 
\subsection{Kernel Functions} \label{sec:kernels}
In this section, we present the library of kernel functions, which were used in building the integral operators. \edit{The candidate kernel library was designed to include both localized and smoothly varying operators capable of capturing different spatial interaction patterns within the input functions. The selection was guided by three principles: (i) Exponential and Radial Basis Function (RBF)-type kernels were primarily selected because their bandwidth parameter provides direct control over the locality of the operator, enabling the framework to capture both short-range and long-range spatial dependencies; (ii) Including both isotropic kernels, which depend on the Euclidean distance between points, and separable kernels acting independently along each spatial coordinate to represent directional dependencies; and (iii) Polynomial kernels to provide additional flexibility for capturing lower-order spatial trends and global variations.

Among the kernel functions included in the library, the RBF kernel, also known as the Gaussian kernel, is one of the most widely used kernels in machine learning and functional analysis~\cite{williams2006gaussian}:}

\begin{equation} 
\bm{\psi} (\mathbf{x}, \bm{\xi} )= \exp(- \frac{ \|  \mathbf{x}- \bm{\xi} \|^2 }{ 2 \sigma^2 } ),
\label{eqn:rbf}
\end{equation}
where the parameter $\sigma$ determines the extent of similarity between samples at locations $\mathbf{x}$ and $\bm{\xi}$. Inspired by the concept of RBF kernels, in this work, various candidates of exponential kernel functions were built for function-to-scalar and function-to-function mapping tasks:
\begin{itemize}
\item function-to-scalar:

\begin{equation}\label{eqn:kernel}
\bm{\psi}(\bm{\xi}) :\quad \exp\left(-\frac{\zeta^2+\eta^2}{\beta}\right), \quad  \exp\left(-\frac{\zeta}{\beta}\right), \quad  \exp\left(-\frac{\eta}{\beta}\right), \
\end{equation}

\item function-to-function:

\begin{align*}
\bm{\psi}(\mathbf{x}, \bm{\xi}) :\quad
& \exp\left(-\frac{({x}-{\zeta})^2+({y}-{\eta})^2}{\beta}\right), \quad
 \exp\!\left(-\frac{\sqrt{(x-\zeta)^2 + (y-\eta)^2}}{\beta}\right), \\[6pt]
& \exp\left(-\frac{({x}-{\zeta})^2}{\beta}\right), \quad
  \exp\left(-\frac{({y}-{\eta})^2}{\beta}\right), \quad
  \exp\left(-\frac{{x}-{\zeta}}{\beta}\right), \quad 
  \exp\left(-\frac{{y}-{\eta}}{\beta}\right),
\end{align*}

\end{itemize}
where $\bm{\xi}: (\zeta,\eta)$ and $\mathbf{x}: (x,y)$ could be perceived as the input function and output function data points, respectively, and $\beta$ is the bandwidth parameter that determines the kernel's spread and locality. The RBF kernel measures the similarity between points based on their Euclidean distance. The kernel value decreases exponentially as the distance increases, meaning that points far apart have negligible influence on each other. A small $\beta$ results in a highly localized function, meaning that only nearby points contribute significantly. Conversely, a large $\beta$ results in a smoother function where more distant points contribute.

\noindent In the case of function-to-function mapping, another kernel function was defined as:

\begin{equation}
\bm{\psi}(\mathbf{x}, \bm{\xi})=
\begin{cases}
0, & \text{if } D \geq 1 \\[6pt]
1, & \text{if } D < 0,
\end{cases}
\end{equation}
which depends on the distance between the coordinates of the input and output domains, $D = \sqrt{\frac{({x}-{\zeta})^2+({y}-{\eta})^2}{\beta}}$. \edit{For each kernel form, the library included multiple candidate bandwidth values $\beta_j$ in order to capture spatial interactions across different length scales~\cite{arzani2024interpreting}.}

\noindent Some other kernel functions that we call polynomial kernels were  additionally used in this study:

\begin{itemize}

\item function-to-scalar: 

\begin{equation}\label{eqn:kernel}
\bm{\psi}(\bm{\xi}) :\quad \zeta ,\quad \eta, \quad  \zeta^2, \quad  \eta^2 ,\quad \zeta\eta,
\end{equation}

\item function-to-function:

\begin{equation}\label{eqn:kernel}
\bm{\psi}(\mathbf{x}, \bm{\xi}) :\quad {x}\zeta+{y}\eta, \quad  ({x}-\zeta)^2, \quad  ({y}-\eta)^2,
\end{equation}

\end{itemize}

It should be noted that in the case of function-to-scalar mapping, we assumed that different regions of the input function contribute uniquely to the scalar output. \edit{To introduce spatial localization into the function-to-scalar mapping, the kernels are shifted across the input domain as:}

\begin{equation} 
\bm{\psi}(\zeta,\eta) = \exp\left(-\frac{{(\zeta - h})^2+({\eta - h})^2}{\beta} \right),
\end{equation}
where the parameter $h$ characterizes the kernel's spatial localization. Similar to the bandwidth, this parameter was varied across the input domain, for instance, in increments of 0.2 to generate different candidate kernels in the library. \edit{The specific step size for $h$ was selected empirically for each test case to balance spatial localization and prediction accuracy.} Introducing kernel-based integral terms, equipped by the parameter $h$, into the library allows each term to represent the significance of a localized input area and captures how the influence of neighboring points diminishes exponentially.

\begin{figure}[h!]
\centering
\includegraphics[scale=.35]{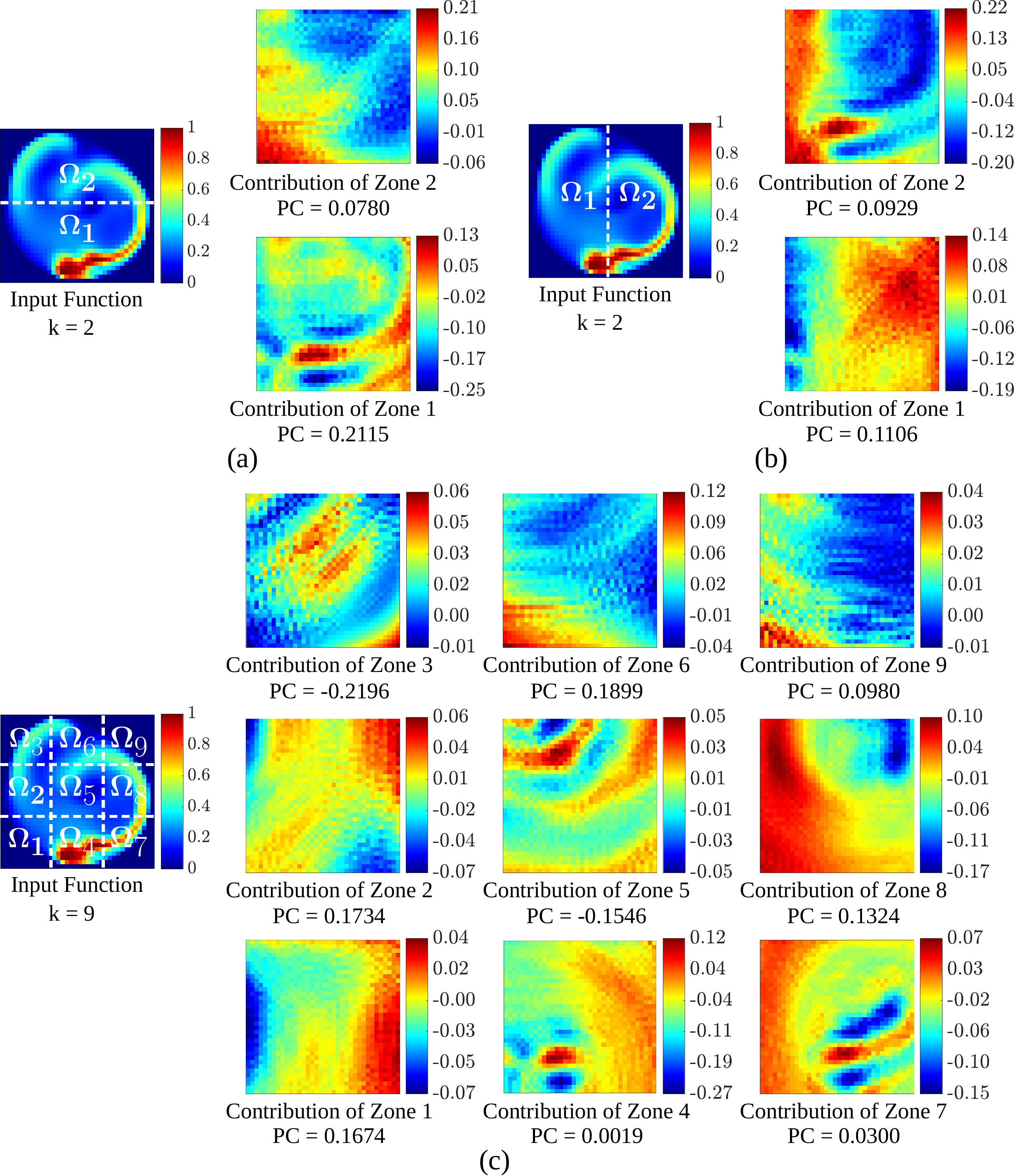}
\caption{\edit{Sensitivity of the zonal decomposition to the number of input subregions $k$ for the function-to-function mapping task in the blood flow test case. (a) Decomposition using $k=2$ with a horizontal partition of the input domain. (b) Decomposition using $k=2$ with a vertical partition. (c) Decomposition using $k=9$ using a $3\times3$ grid partition. For each configuration, representative mean-centered zonal contributions are shown together with the corresponding Pearson correlation coefficients (PC) relative to the full XAI model prediction.}}
\label{fig:diff_k}
\end{figure}

\edit{
\subsection{Sensitivity to the Number of Input Subregions ($k$)} \label{sec:app_zones}

To assess the sensitivity of the zonal decomposition to the choice of $k$, we applied the framework to the function-to-function mapping in the blood flow test case using $k=2$ (with both horizontal and vertical partitions) and $k=9$ ($3\times3$ grid), in addition to the $k=4$ ($2\times2$ grid) used throughout the main text. Fig.~\ref{fig:diff_k} presents the mean-centered zonal contributions and corresponding Pearson correlation coefficients (PC) for a representative time step (Sample 1 marked on Fig.~\ref{fig:F2F_ZonesCont}a).

A consistent spatial pattern in the output contributions is observed across all values of $k$. For $k=2$ with a horizontal partition (Fig.~\ref{fig:diff_k}a), Zone 1 (lower half) produces a contribution with a spatial pattern that closely resembles the one obtained from Zone 2 (right half), when the same domain is partitioned vertically (Fig.~\ref{fig:diff_k}b). This indicates that the underlying spatial signal occupies the lower-right portion of the input domain and is detected regardless of how the domain is split. With $k=4$ (main text results), this same pattern is captured  by Zone 3 (lower-right quadrant), which corresponds to the intersection of the two $k=2$ zones. This is consistent with the results reported in the main text, where Zone 3 was identified as the most influential region for the WSS prediction.

For $k=9$ (Fig.~\ref{fig:diff_k}c), the finer partition distributes the signal across multiple zones, with Zones 4 and especially Zone 7 producing the spatial pattern most similar to that observed in Zone 3 of the $k=4$ decomposition. These zones occupy the bottom-center and bottom-right portions of the domain, overlapping directly with the region identified as dominant at coarser resolutions.

These results support three conclusions. First, the dominant spatial pattern identified by the framework is robust to the choice of $k$ where the same physical region emerges as the primary driver of the output regardless of partition resolution. Second, $k=2$ is too coarse to isolate this region as it mixes the relevant features with a large portion of uninformative input, diluting the explanation. Third, $k=9$ confirms the same finding but spreads the signal across several adjacent zones, making the explanation harder to interpret without providing additional physical insight. The $k=4$ partition strikes a practical balance as it is fine enough to localize the dominant input region in a single zone yet coarse enough to yield  clearly distinguishable and interpretable contributions.

A similar observation holds for the Pearson correlation values at different resolutions. In the $k=4$ analysis (Fig.~\ref{fig:F2F_ZonesCont}), Zone 3's Pearson correlation at individual time steps can be modest (e.g., PC $= 0.0520$ at S2), yet its average correlation across the full dataset is the highest among all zones. Similarly, at $k=9$, the zones producing the dominant spatial pattern (Zones 4 and 7) show low individual correlations, but the spatial structure they capture remains physically coherent and consistent with the patterns identified at coarser partitions. 


}

\edit{
\subsection{Gradient-weighted Class Activation Mapping (Grad-CAM) Analysis} 
\label{sec:gradcam}

Grad-CAM was originally developed for classification tasks, where gradients are computed with respect to a specific output class~\cite{selvaraju2020grad}. In the present function-to-function mapping problem, however, the U-Net produces a dense $36 \times 36$ output field rather than a single scalar prediction. To adapt Grad-CAM to this setting and maintain consistency with the other explainability analyses, a scalar target was defined by summing all predicted output values $\mathbf{u}_{ij}({x}, {y})$:

\begin{equation} 
{u} = \sum_{ij}\mathbf{u}_{ij}({x}, {y}).
\end{equation}

Gradients of this scalar quantity with respect to the activations of a selected convolutional layer were then computed through backpropagation. The gradients were globally averaged across the spatial dimensions to obtain channel-wise importance weights, which were subsequently combined with the corresponding feature maps to generate a class-agnostic saliency map. A Rectified Linear Unit (ReLU) operation was applied to retain only positive contributions, after which the heatmap was upsampled to the original input resolution and normalized to the range $[0, 1]$ for visualization.

To analyze the influence of representation depth on the explanations, Grad-CAM was separately computed using the encoder layers (Stages 1--4). Fig.~\ref{fig:gradcam} presents Grad-CAM activation maps computed from different encoder stages of the trained U-Net model for representative samples from the function-to-function mapping task. The first row shows the input velocity magnitude fields, while the remaining rows visualize the Grad-CAM saliency maps obtained from encoder stages 1--4.

The shallow encoder stages produce highly localized activation patterns that closely follow fine-scale structures and sharp feature boundaries in the input field. In particular, Encoder Stage~1 strongly highlights regions with steep spatial gradients, especially the transition between low and high velocity magnitudes in the lower-right portion of the domain (corresponding to Zone~3 in the zonal decomposition framework). As the encoder depth increases, the spatial resolution of the activation maps decreases due to successive pooling operations and larger receptive fields. Encoder Stage~2 begins to isolate more compact and abstract regions of model sensitivity, while Encoder Stage~3 captures broader spatial structures associated with regional flow organization rather than localized texture. At the bottleneck representation (Encoder Stage~4), the saliency maps become highly coarse and spatially aggregated, emphasizing only large-scale domain-level distinctions. This progressive transition from localized gradient-sensitive features to spatially coarse semantic representations reflects the hierarchical feature abstraction learned by the encoder.

Importantly, the salient regions emphasized by Grad-CAM for Stage 1 remain qualitatively consistent with the influential regions identified by the proposed XAI framework as well as the zone-based Kernel SHAP and occlusion sensitivity analyses. Across all methods, regions containing strong local gradients consistently emerge as the most influential contributors to the prediction. 

\begin{figure}[h!]
\centering
\includegraphics[scale=.25]{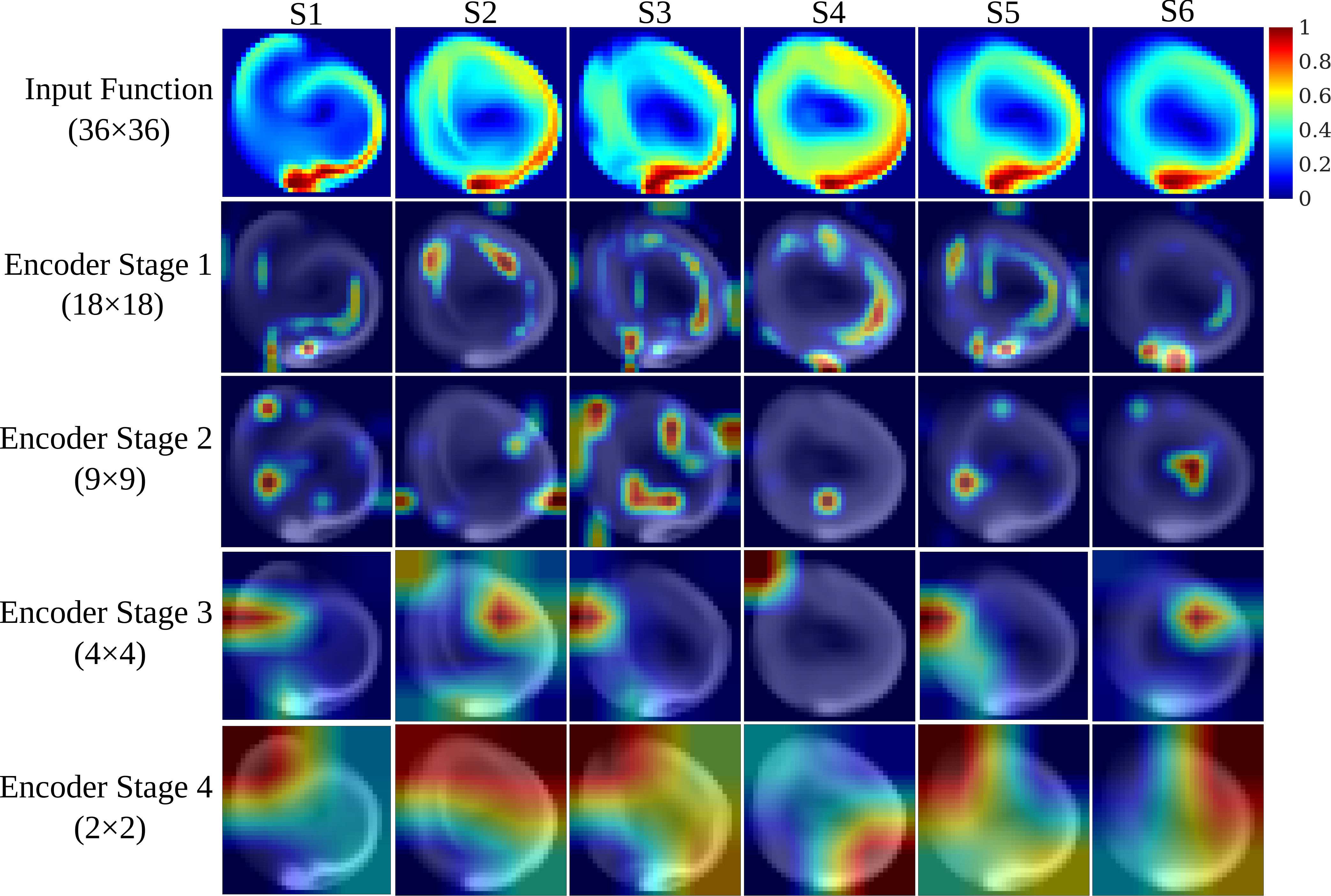}
\caption{\edit{Grad-CAM heatmaps for the function-to-function mapping task in the blood flow test case, computed at four encoder stages of the U-Net. The top row shows the input velocity fields for six representative test samples (S1--S6). Subsequent rows display the Grad-CAM saliency maps at Encoder Stages 1--4, with spatial resolutions of $18 \times 18$, $9 \times 9$, $4 \times 4$, and $2 \times 2$, respectively. All heatmaps are upsampled to the original $36 \times 36$ resolution and normalized to $[0, 1]$. Warmer colors indicate regions with higher importance for the prediction.}}
\label{fig:gradcam}
\end{figure}
}

\edit{
\subsection{Data Preprocessing}
\label{sec:preprocessing}

The blood flow dataset used in this study was processed with Visualization Toolkit (VTK) files. A planar slice passing through the aneurysm region was first extracted from the three-dimensional simulation domain. The velocity magnitude field evaluated on this plane was used as the input function to the XAI model. For the output function in the function-to-function mapping task, the wall shear stress (WSS) magnitude corresponding to the vessel wall intersecting the aneurysm region was projected onto a two-dimensional plane parallel to the velocity slice. The projected WSS fields and corresponding two-dimensional spatial coordinates were extracted from the projected aneurysm plane.

For the global velocity-magnitude preprocessing, the scattered field data were interpolated with linear interpolation onto a uniform Cartesian grid spanning the full normalized domain $[0,1]\times[0,1]$ and grid of size $36\times36$. For the localized WSS output fields used in the function-to-function mapping task, a smaller region of interest was extracted from the projected wall field using the coordinate ranges:

\begin{equation}
0.45 < x < 0.90,
\qquad
0.05 < y < 0.50.
\end{equation}

Only scattered data points located within this region were retained. The extracted local field was then interpolated onto a regular Cartesian grid using linear interpolation. The localized WSS fields were resampled onto structured grids of size $36\times36$, where the grid points were generated uniformly over the selected spatial region. The resulting structured datasets (interpolated fields) were used for subsequent operator-learning analysis. 

}

\edit{
\subsection{Model Sensitivity and Robustness Analysis} 
\label{sec:sensitivity}

To assess the robustness of the proposed framework, we conducted a sensitivity analysis on the Max WSS function-to-scalar mapping task in the blood flow test case. The analysis examines the effect of varying the kernel bandwidth range ($\beta$) and the regularization parameter ($\lambda$) on both prediction accuracy and the zonal contribution patterns, as well as the stability of the explanations under input noise perturbations.

Fig.~\ref{fig:sensitivity}a shows the results when the lower bound of the bandwidth range is fixed at $\beta_{\min} = 0.1$ and the upper bound $\beta_{\max}$ is varied alongside $\lambda$. The regularization parameter $\lambda$ was selected through empirical tuning to balance prediction accuracy and numerical stability of the regression problem. The heatmaps display the Mean absolute error (AE) (left) and Max AE (center) across the hyperparameter grid. Lower values of $\lambda$ and moderate values of $\beta_{\max}$ yield the lowest errors, while excessively large bandwidths or strong regularization degrade accuracy. The box plot (right) shows the distribution of Pearson correlation coefficients across all hyperparameter combinations for each zone. The prediction errors are more sensitive to the regularization parameter than $\beta_{\max}$, whereas the ranking of the zonal contributions remains relatively stable across the tested parameter range. In particular, Zone~3 consistently exhibits the highest correlation with the XAI model prediction.

Fig.~\ref{fig:sensitivity}b presents the complementary experiment, where $\beta_{\max}$ is fixed at 2.5 and $\beta_{\min}$ is varied alongside $\lambda$. Since the input functions are discretized on a $36 \times 36$ grid with coordinates specified by $[0,\,1]$, the grid spacing is $\Delta = 1/35 \approx 0.028$. The candidate values of $\beta_{\min}$ are chosen as multiples of this grid spacing, ranging from approximately $1\Delta$ ($\beta_{\min} = 0.03$) to $30\Delta$ ($\beta_{\min} = 0.80$). Setting $\beta_{\min}$ too close to the grid spacing leads to poorly resolved kernel evaluations and numerically unstable configurations, as indicated by the cells marked N/A in the heatmaps. These instabilities arise because kernels with bandwidths near or below the grid resolution produce poor integral evaluations. In contrast, setting $\beta_{\min}$ too large prevents the model from capturing fine-scale spatial features, which also degrades accuracy. The selected baseline value of $\beta_{\min} = 0.1 \approx 3$--$4\Delta$ balances these considerations, lying in a region where the model achieves low prediction error and stable zonal explanations. The same overall pattern observed in Fig.~\ref{fig:sensitivity}a holds: prediction errors increase with larger $\lambda$ or narrower bandwidth ranges, but the zonal ranking remains consistent, with Zone~3 maintaining the highest median correlation across all valid configurations.

Fig.~\ref{fig:sensitivity}c examines the effect of additive noise on the input velocity fields, with noise levels ranging from 0\% to 20\% of the input magnitude. The output quantity is kept unchanged. The left panel shows that both Mean AE and Max AE increase gradually with noise level, as expected, but the model remains reasonably accurate up to 5\% noise. The right panel tracks the Pearson correlation of each zone's contribution as a function of noise level. At low noise levels (0--5\%), Zone~3 maintains the highest correlation, consistent with the baseline results reported in the main text. At higher noise levels (10--20\%), the relative ranking among zones shifts, with Zone~4 overtaking Zone~3 as the most correlated region. This suggests that while the framework's explanations are stable under moderate input corruption, stronger noise can alter the balance among zones. 

\begin{figure}[h!]
\centering
\includegraphics[scale=.45]{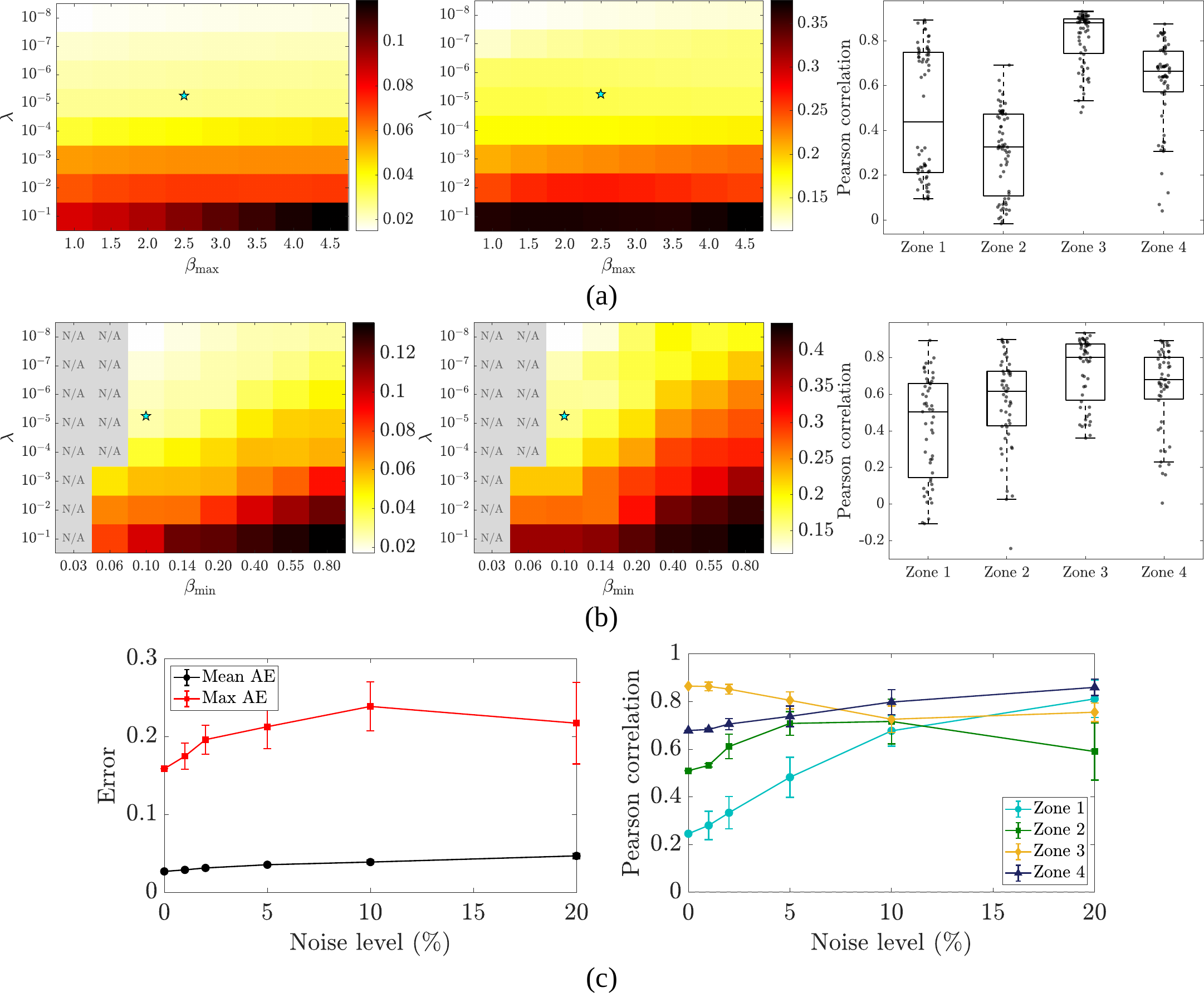}
\caption{\edit{Sensitivity analysis for the Max WSS function-to-scalar mapping in the blood flow test case. (a) Effect of varying $\beta_{\max}$ and $\lambda$ with $\beta_{\min}$ fixed at 0.1: Mean AE (left), Max AE (center), and distribution of zonal Pearson correlations across all configurations (right). The star marks the baseline setting used in the main text. (b) Effect of varying $\beta_{\min}$ and $\lambda$ with $\beta_{\max}$ fixed at 2.5. N/A cells indicate unstable configurations. (c) Effect of input noise perturbation: prediction errors (left) and zonal Pearson correlations (right) as a function of noise level (0--20\%). Error bars denote standard deviation across five random seeds.}}
\label{fig:sensitivity}
\end{figure}
}

\edit{
\subsection{Fourier Neural Operator Baseline for Accuracy Comparison} 
\label{sec:fno}

FNO is a well-established neural operator architecture that learns mappings between function spaces through spectral convolution in the Fourier domain. We implemented the models using the NeuralOperator library~\cite{kossaifi2024library} and trained them with the Adam optimizer (learning rate of $10^{-3}$ and weight decay of $10^{-5}$), along with a cosine annealing scheduler ($T_{\max} = 2000$, $\eta_{\min} = 10^{-5}$). Each model was trained for up to 2000 epochs, with early stopping based on the lowest test relative $L^2$ error. To ensure a fair comparison, all FNO models were trained using the same datasets, train/test splits, and normalization procedures as the proposed XAI model. Every experiment was run with five different random seeds. Model performance is assessed using the Mean AE and Max AE, consistent with the evaluation metrics used for the XAI model in Table~\ref{tab:accuracy}.

For the function-to-function mapping in the blood flow test case, we used the standard FNO architecture without modification. The model takes a single-channel $36 \times 36$ velocity magnitude field as input and produces a single-channel $36 \times 36$ output corresponding to the localized high-WSS region. The model consists of 4 Fourier layers with 64 hidden channels and retains $16 \times 16$ Fourier modes. The input, augmented with a grid-based positional embedding, first passes through a lifting layer into a higher-dimensional feature space. It then proceeds through a sequence of spectral convolution layers with Gaussian Error Linear Unit (GELU) activations and channel multilayer perceptron (MLP) blocks equipped with soft-gating skip connections, before a final projection layer maps the features back to the output space.

Because FNO is inherently designed for function-to-function mappings, we adapted it for scalar output prediction by attaching a fully connected (FC) regression head to the FNO backbone. Two variants were designed to accommodate the different input resolutions of the test cases. For the blood flow cases ($36 \times 36$ input functions), the FNO output feature maps were flattened and passed through a three-layer FC network ($1296 \rightarrow 512 \rightarrow 128 \rightarrow 1$) with GELU activations and dropout ($p = 0.1$); the backbone retained the same configuration as in the function-to-function task (4 layers, 64 hidden channels, $16\times16$ Fourier modes). For the airfoil cases ($100 \times 100$ input resolution), directly flattening the full feature maps would result in a prohibitively large input dimension for the FC layers. Instead, we applied multi-statistic pooling, extracting the mean, maximum, and standard deviation over the spatial dimensions of each channel to form a compact 192-dimensional feature vector ($64 \times 3$ features). This vector was then passed through a three-layer FC head ($192 \rightarrow 256 \rightarrow 64 \rightarrow 1$) with GELU activations and dropout ($p = 0.1$). To better capture the higher-resolution spatial structure, the number of retained Fourier modes is increased to $32 \times 32$.}

\edit{
\subsection{Sum of Contributions} 
\label{sec:sum_contributions}
Fig.~\ref{fig:Sum_zones_blood} examines how combinations of zonal contributions reproduce the overall XAI model prediction for the aneurysm test case. The first column in each subfigure shows the best two-zone combination, and the second column shows the best three-zone combination. Pearson correlation coefficients (PC) are computed between each summed contribution and the total XAI model prediction. Notably, for all three quantities --- maximum WSS (Fig.~\ref{fig:Sum_zones_blood}a), spatially averaged WSS (Fig.~\ref{fig:Sum_zones_blood}b), and the spatial average of the divergence of normalized WSS (Fig.~\ref{fig:Sum_zones_blood}c) --- the combination of Zones~2 and~4 provides the best two-zone approximation, with correlations of PC $= 0.9056$, $0.9464$, and $0.9387$, respectively. This suggests that the information contained in these two regions alone is sufficient to capture the dominant temporal behavior of the system across all three WSS-related metrics. Adding a third zone further improves the approximation: Zones~1, 2, and~4 (PC $= 0.9855$) for maximum WSS; Zones~2, 3, and~4 (PC $= 0.9898$) for spatially averaged WSS; and Zones~1, 2, and~3 (PC $= 0.9654$) for the spatial average of the divergence of normalized WSS. Overall, these results suggest that although certain zones dominate the prediction individually, the temporal behavior of the output is distributed across multiple interacting regions. 

\begin{figure}[h!]
\centering
\includegraphics[scale=.22]{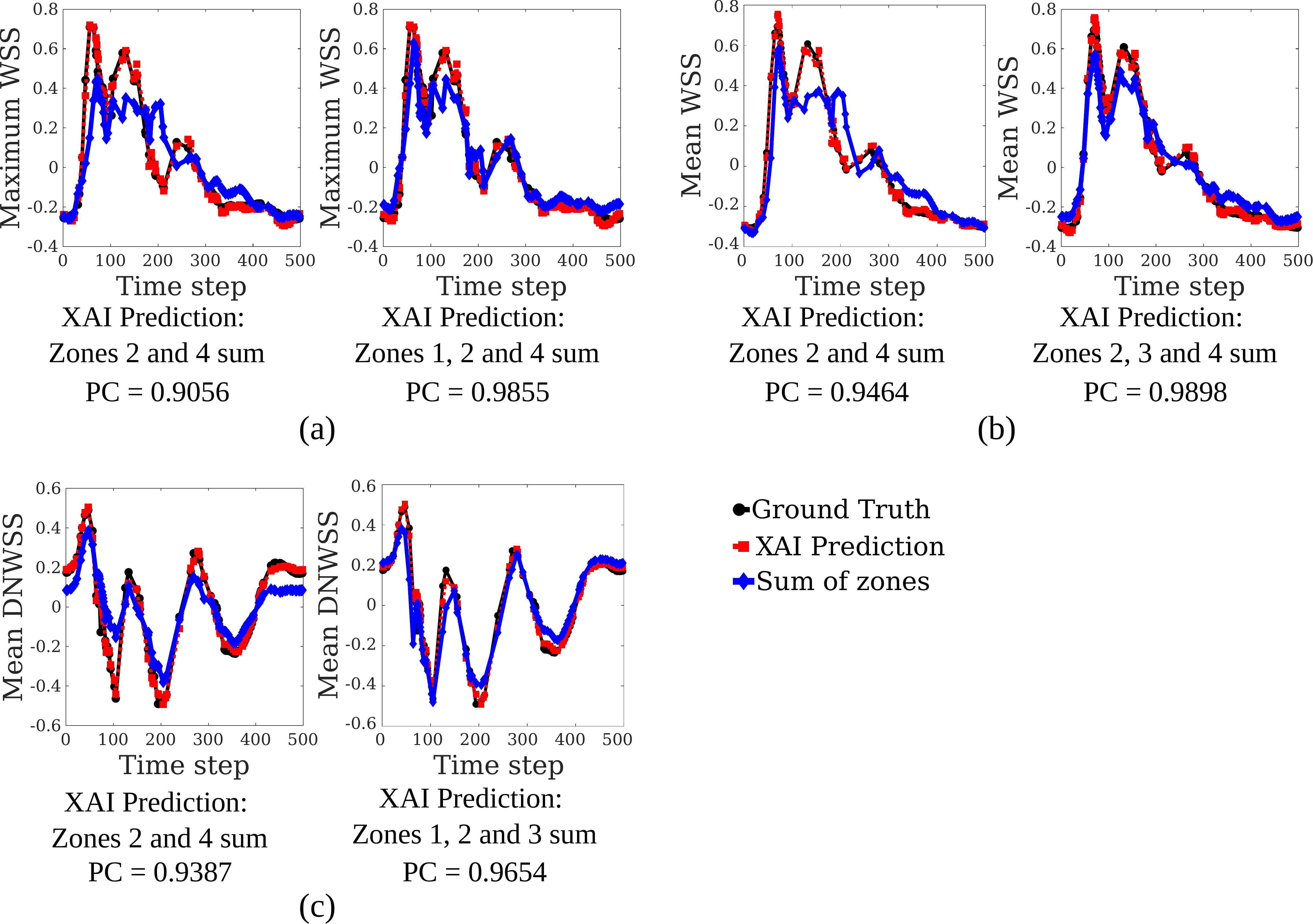}
\caption{\edit{Combinations of zonal contributions for the three function-to-scalar mapping tasks in the blood flow test case: (a) maximum WSS, (b) spatially averaged WSS (Mean WSS), and (c) spatial average of the divergence of normalized WSS (Mean DNWSS). The first column shows the best two-zone combination (blue), and the second column shows the best three-zone combination (blue), alongside the mean-centered ground truth (black) and XAI model prediction (red). Pearson correlation coefficients (PC) are indicated for each combination.}}
\label{fig:Sum_zones_blood}
\end{figure}

For the airfoil test case, combinations of zonal contributions were also evaluated. Fig.~\ref{fig:Sum_zones_airfoil} summarizes the best-performing two-zone and three-zone combinations for the drag and lift coefficient predictions at both pitching frequencies. For the lower-frequency case ($f_p = 0.05$), the drag coefficient is reproduced most accurately by combining the contributions from Zones~1 and~2, yielding a strong agreement with the overall XAI model prediction (PC $= 0.9873$). Adding Zone~4 slightly reduces the correlation (PC $= 0.9795$), indicating that the dominant drag dynamics at this frequency are already captured primarily by the lower-left and upper-left regions of the domain. A similar trend is observed for the lift coefficient, where the combination of Zones~2 and~4 produces a strong approximation of the XAI model prediction (PC $= 0.9716$). Including Zone~1 further improves the agreement (PC $= 0.9870$), suggesting that the lift dynamics arise from interacting contributions between the upper flow structures and additional lower-domain features. At the higher pitching frequency ($f_p = 0.2$), the oscillatory dynamics become more complex, and the dominant combinations shift toward wake-dominated regions. For both drag and lift predictions, the combination of Zones~3 and~4 provides the best two-zone approximation (PC $= 0.8869$ for drag and PC $= 0.8378$ for lift), highlighting the increasing influence of wake and vortex-shedding structures at higher oscillation frequencies. The best-performing three-zone combinations are Zones~1, 2, and~4 for the drag coefficient (PC $= 0.8468$) and Zones~1, 3, and~4 for the lift coefficient (PC $= 0.8591$). 

Overall, these results suggest that the aerodynamic response at lower pitching frequency can be explained reasonably well using a limited subset of localized flow regions, whereas the higher-frequency case requires contributions from broader interacting wake structures to reproduce the oscillatory dynamics. The additive structure of the proposed framework makes it possible to explicitly identify and quantify these distributed spatial interactions.

\begin{figure}[h!]
\centering
\includegraphics[scale=.21]{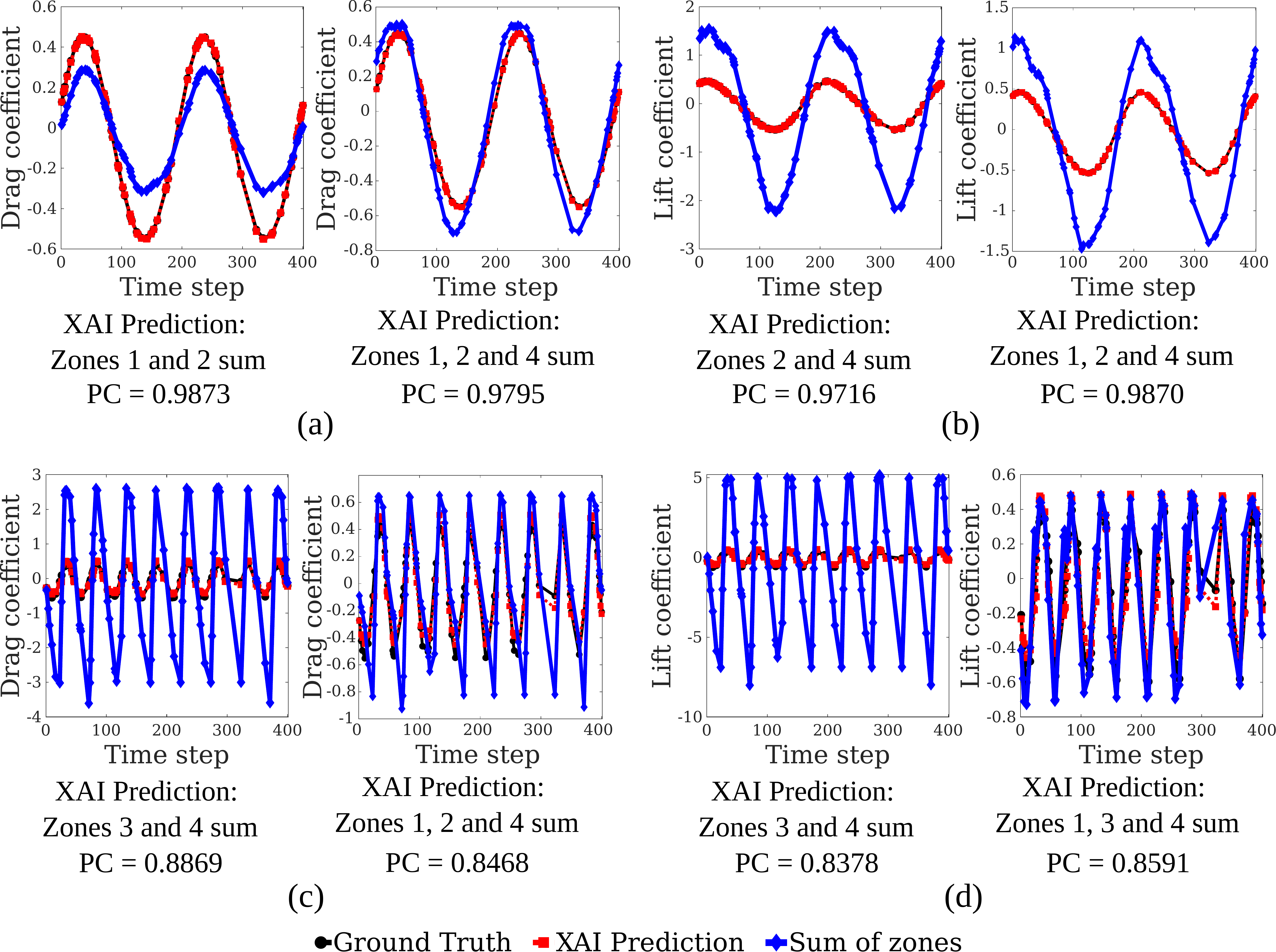}
\caption{\edit{Combinations of zonal contributions for the airfoil function-to-scalar mapping tasks. (a) Drag coefficient prediction at nondimensional pitching frequency $f_p = 0.05$. (b) Lift coefficient prediction at $f_p = 0.05$. (c) Drag coefficient prediction at $f_p = 0.2$. (d) Lift coefficient prediction at $f_p = 0.2$. In each subfigure, the left panel shows the best-performing two-zone combination and the right panel shows the best-performing three-zone combination. The summed zonal contributions (blue) are plotted along with the mean-centered ground truth data (black) and the XAI model prediction (red). Pearson correlation coefficients (PC) quantify the agreement between the summed zonal contributions and the XAI model prediction.}}
\label{fig:Sum_zones_airfoil}
\end{figure}

}

\edit{
\subsection{Role of Non-linearity}
\label{sec:nonlinear}

\edit{To further investigate the role of nonlinearities in the proposed framework, we conducted an additional experiment on the function-to-function mapping task by augmenting the candidate library with nonlinear integral terms. Specifically, nonlinear activation functions, including exponential, hyperbolic tangent, and quadratic mappings, were applied to the outputs of selected integral operators, increasing the total number of candidate terms in the library from 2405 to 3609. This experiment was conducted to compare the accuracy, and we should highlight that adding nonlinearity to the output of the integrals breaks the linear spatial decomposition utilized for the XAI analysis. For the same test case and random seed, the nonlinear operator achieved a slightly lower Mean AE = $0.0690$ versus $0.0734$ for the linear operator, while the Max AE increased (Max AE = $0.6215$ versus $0.5444$). 

The limited improvement in predictive accuracy despite the increased expressive capability of the nonlinear model appears to be associated with severe numerical ill-conditioning of the candidate library, which is a known challenge in library-based regression~\cite{feng2026ill} where adding more features can increase the correlation between the columns and adversely impact the solution to the linear system of equations. 


}

}

\edit{
\subsection{Time-Varying Analysis: Function-to-Function Mapping}
\label{sec:f2f_timevarying}

To further investigate the zonal contributions in the function-to-function mapping across the entire dataset, spatially averaged WSS values are computed for the ground truth data and the integral operator's output for each zone. These averaged values convert the function outputs into scalar measures at each time step, allowing for time-varying analysis. Fig.~\ref{fig:F2F2S} presents the results of this time-varying analysis. The first two panels show the mean-centered ground truth and the mean-centered XAI model prediction, respectively; their close agreement confirms that the model captures the dominant temporal trends in the spatially averaged WSS with reasonable accuracy. The next four panels display the contribution of each individual zone, along with the mean-centered ground truth data and XAI model predictions. 

The qualitative and quantitative comparison between the zonal contributions and the XAI model prediction shows that the averaged contribution from Zone~3 exhibits the highest correlation among all zones (PC $= 0.6888$) and aligns most closely with the full XAI model prediction across all time steps. This finding is consistent with our previous observations obtained by evaluating the functional outputs at selected time steps. The remaining zones show weaker or negative correlations, with Zone~2 contributing minimally (PC $= -0.1476$). The last two panels explore combinations of zonal contributions: adding Zone~1 to Zone~3 (PC $= 0.6888$) marginally improves the match, while the three-zone combination of Zones~1, 3, and~4 (PC $= 0.8419$) provides the closest approximation to the full XAI model prediction. Nevertheless, Zone~3 remains the dominant contributor, underscoring its pivotal role in shaping the WSS pattern within the integral operator model.

\begin{figure}[h!]
\centering
\includegraphics[scale=.16]{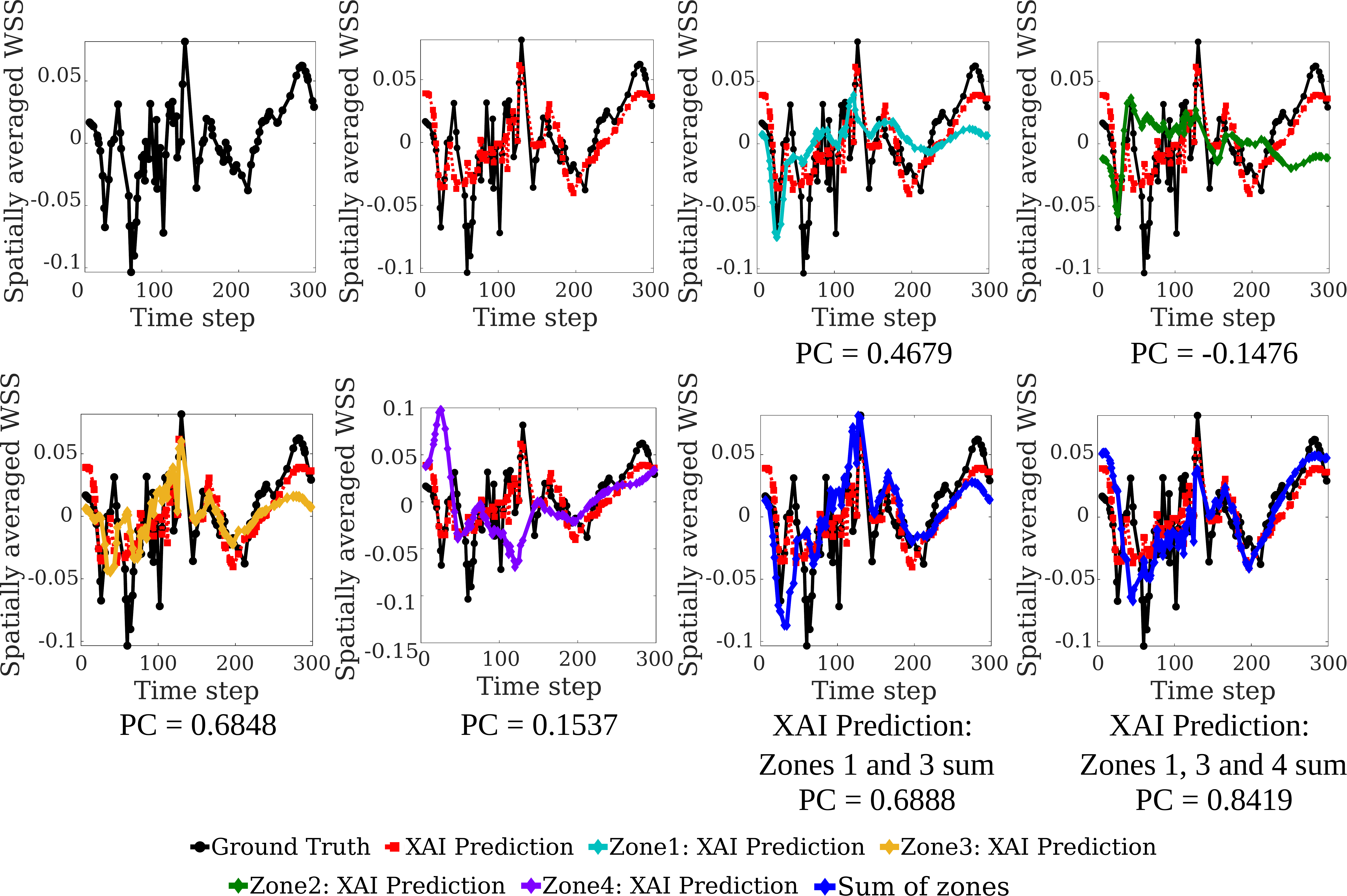}
\caption{\edit{Time-varying analysis of zonal contributions for the function-to-function mapping task in the blood flow test case. Spatially averaged WSS values are computed from the output fields at each test time step. The first two panels show the mean-centered ground truth and XAI model prediction. The next four panels show the mean-centered contribution of each zone, with Pearson correlation coefficients (PC) indicated. The last two panels show the best two-zone and three-zone combinations alongside the mean-centered ground truth and XAI prediction.}}
\label{fig:F2F2S}
\end{figure}
}

\edit{
\subsection{Additional Experiment: Semi-linear Elliptic PDE}
\label{sec:newdataset}

We further assess the applicability of the proposed framework and demonstrate a case where the method does not work as accurately as our presented main test cases, by considering irregular input functions (without dominant coherent patterns). We applied the methodology to a synthetic dataset generated from a  semi-linear elliptic PDE problem of the form:

\begin{equation}
-\Delta u + u^2 = f,
\end{equation}
following the synthetic data-generation procedure described in~\cite{hasani2025generating}. Here, $f$ represents the input forcing term and $u$ denotes the corresponding solution field defined on a two-dimensional spatial domain with zero Neumann boundary conditions, as explained in~\cite{hasani2025generating}. Both the input and output functions were discretized on uniform $36\times36$ Cartesian grids. In this dataset, the input functions exhibit spatially chaotic patterns (challenging XAI modeling), while the corresponding output functions remain comparatively smooth.

Fig.~\ref{fig:newdataset} presents representative results of the XAI model for selected samples from the test dataset. For each sample, the input function, ground truth output, and corresponding XAI model prediction are shown. In addition, the four zonal contributions (mean-centered predictions associated with each input subregion) together with their summed contribution are visualized for each test case. As shown in Fig.~\ref{fig:newdataset}, the proposed XAI model exhibits lower predictive accuracy for this dataset (Mean AE = $0.1084$ and Max AE = $0.6759$ on the test dataset) compared with the other function-to-function test case considered in this study. In contrast to the previous examples, however, the input functions in this dataset display irregular patterns without clearly identifiable coherent structures. Consequently, the resulting zonal contributions become less stable and more difficult to interpret physically, with individual regions rarely exhibiting clear or consistent dominant behavior across samples.

More generally, these results suggest that the current linear decomposition framework performs most effectively when the input functions contain spatially organized or coherent structures that can be associated with localized regions of the domain. In cases where the input patterns are highly irregular or weakly structured, the resulting decomposed contributions may become less accurate even if portions of the overall output behavior are still captured by the model.

\begin{figure}[h!]
\centering
\includegraphics[scale=.2]{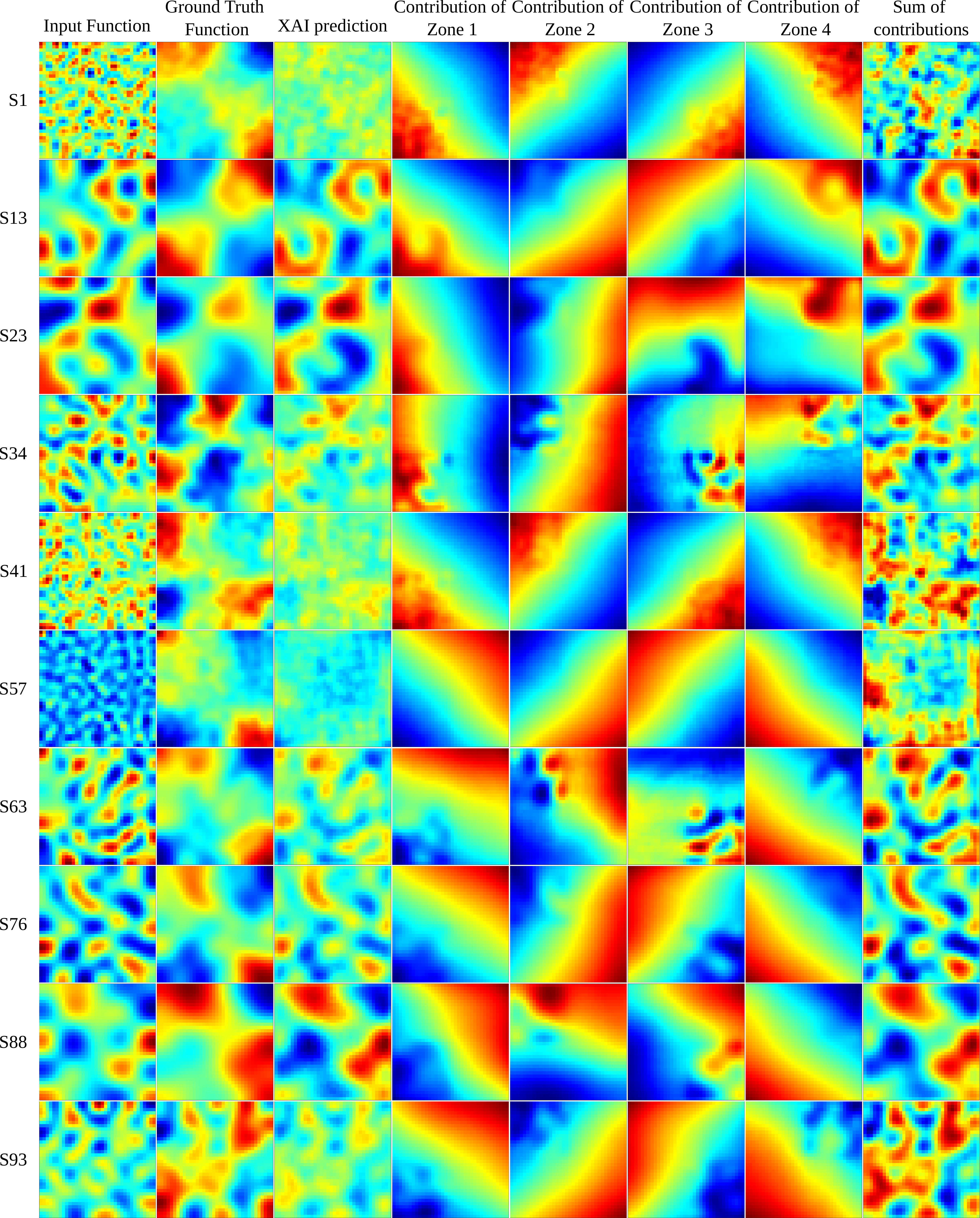}
\caption{\edit{Results of the additional function-to-function mapping experiment using the synthetic semi-linear elliptic PDE dataset. The first column shows representative input functions for selected test samples. The second and third columns show the corresponding ground truth outputs and XAI model predictions, respectively. Columns 4--7 present the mean-centered zonal contributions obtained from the four input subregions, while the final column shows the mean-centered sum of contributions.}}
\label{fig:newdataset}
\end{figure}
}

\edit{
\subsection{Computational Complexity} 
\label{sec:complexity}

All integrations were performed on uniform Cartesian grids, and the integral operators were evaluated numerically using the trapezoidal rule over the discretized spatial domains. During the post-processing zonal decomposition analysis, special care was taken to ensure consistent treatment of the interfaces between adjacent zones. Because the trapezoidal rule assigns nonzero weights to boundary points, these shared interface contributions would otherwise be omitted when integrating over individual subdomains separately. To preserve the additive consistency of the overall integral evaluation, additional integrations were therefore performed along the shared boundaries, and the resulting contributions were distributed appropriately among the neighboring zones.

\noindent For the function-to-function mapping task, let:
\begin{itemize}
    \item $N \times N$ denote the spatial resolution of the discretized field,
    \item $N_s$ denote the number of training samples,
    \item $N_k$ denote the number of kernel functions,
    \item $N_b$ denote the number of bandwidth values associated with each kernel,
    \item $N_t$ denote the total number of candidate integral terms in the library.
\end{itemize}

\noindent The total number of candidate integral terms scales approximately as:
\begin{equation}
N_t \sim N_k N_b,
\end{equation}
with additional contributions arising from lifted transformations of the input functions.

The dominant computational cost arises during the construction of the feature library matrix. For each training sample, the framework evaluates a two-dimensional integral for every candidate term at each output spatial location. Using the trapezoidal rule over uniform Cartesian grids, the resulting computational complexity scales approximately as:

\begin{equation}
\mathcal{O}\left(N_s N_t N^4\right),
\end{equation}
which constitutes the primary computational bottleneck of the framework. Once the feature matrix is constructed, training reduces to solving a regularized linear system, described in the main text. The computational computational complexity of the solver is related to $N_t$ and the choice of the linear solver. It should be noted that unlike neural-network-based operator-learning methods, the proposed framework does not require iterative backpropagation or multi-epoch optimization.

After training, inference is computationally less expensive, since predictions only require evaluating a linear combination of the learned integral terms:

\begin{equation}
\mathbf{u} = \sum_{m=1}^{N_t} w_m I_m,
\end{equation}
leading to an inference complexity related to $N_t$.


All computations were performed on a computer equipped with an Intel Xeon W-2245 CPU (8 cores, 16 threads, 3.90 GHz). The framework was implemented in MATLAB without parallelization or GPU acceleration. For representative cases, the total runtime was approximately 5.4 minutes for the Max WSS function-to-scalar mapping task, 2.1 minutes for the airfoil drag prediction task at $f_p=0.2$, and 611.6 minutes for the function-to-function mapping task in the blood flow case. The substantially higher runtime for the function-to-function mapping reflects the increased cost of evaluating two-dimensional integral operators at every output spatial location during feature-library construction.
}

\clearpage

\printbibliography

\end{document}